\documentclass[article]{IEEEtran}
\usepackage{cite}
\usepackage{color}
\usepackage[pdftex]{graphicx}
\graphicspath{{fig/}{jpeg/}}
\usepackage[cmex10]{amsmath}
\usepackage{amssymb}
\usepackage{algorithm}
\usepackage{algpseudocode}
\usepackage{amsmath,amssymb,lipsum}
\usepackage{hyperref}
\usepackage{comment}
\usepackage{graphicx}
\usepackage[font=small]{caption}
\usepackage{subcaption}
\input{mysymbol.sty}
\usepackage{needspace}

\usepackage{theorem}
\newtheorem{thm}{Theorem}
\newtheorem{lemma}{Lemma}
\newtheorem{proposition}{Proposition}
\theoremstyle{definition}
\newtheorem{assumption}{Assumption}
\newtheorem{remark}{Remark}

\title{RES: Regularized Stochastic BFGS Algorithm}
\author{Aryan Mokhtari and Alejandro Ribeiro 
\thanks{Work in this paper is supported by ARO W911NF-10-1-0388, NSF CAREER CCF-0952867, and ONR N00014-12-1-0997. The authors are with the Department of Electrical and Systems Engineering, University of Pennsylvania, 200 South 33rd Street, Philadelphia, PA 19104. Email: \{aryanm, aribeiro\}@seas.upenn.edu. Part of the results in this paper appeared in \cite{cMokhtariRibeiro13} and \cite{cMokhtariRibeiro14}.}}

\begin{document}

\maketitle
\thispagestyle{empty}
\pagestyle{empty}

\begin{abstract}
RES, a regularized stochastic version of the Broyden-Fletcher-Goldfarb-Shanno (BFGS) quasi-Newton method is proposed to solve convex optimization problems with stochastic objectives. The use of stochastic gradient descent algorithms is widespread, but the number of iterations required to approximate optimal arguments can be prohibitive in high dimensional problems. Application of second order methods, on the other hand, is impracticable because computation of objective function Hessian inverses incurs excessive computational cost. BFGS modifies gradient descent by introducing a Hessian approximation matrix  computed from finite gradient differences. RES utilizes stochastic gradients in lieu of deterministic gradients for both, the determination of descent directions and the approximation of the objective function's curvature. Since stochastic gradients can be computed at manageable computational cost RES is realizable and retains the convergence rate advantages of its deterministic counterparts. Convergence results show that lower and upper bounds on the Hessian egeinvalues of the sample functions are sufficient to guarantee convergence to optimal arguments. Numerical experiments showcase reductions in convergence time relative to stochastic gradient descent algorithms and non-regularized stochastic versions of BFGS. An application of RES to the implementation of support vector machines is developed.
\end{abstract}

\section{Introduction}\label{sec_Introduction}

Stochastic optimization algorithms are used to solve the problem of optimizing an objective function over a set of feasible values in situations where the objective function is defined as an expectation over a set of random functions. In particular, consider an optimization variable $\bbw \in \reals^{n}$ and a random variable $\bbtheta \in \Theta\subseteq\reals^p$ that determines the choice of a function $f(\bbw,{\bbtheta}):\reals^{n\times p}\to\reals$. The stochastic optimization problems considered in this paper entail determination of the argument $\bbw^*$ that minimizes the expected value $F(\bbw):=\mbE_{\bbtheta}[f(\bbw,{\bbtheta})]$,
\begin{equation}\label{optimization_problem}
   \bbw^* \ :=\ \argmin_\bbw \mbE_{\bbtheta}[f(\bbw,{\bbtheta})]
          \ :=\ \argmin_\bbw {F(\bbw)}.
\end{equation} 
We refer to $f(\bbw,{\bbtheta})$ as the random or instantaneous functions and to $F(\bbw):=\mbE_{\bbtheta}[f(\bbw,{\bbtheta})]$ as the average function. Problems having the form in \eqref{optimization_problem} are common in machine learning \cite{BottouCun, Bottou, SS} as well as in optimal resource allocation in wireless systems \cite{AryanAle, Ribeiro10, Ribeiro12}.

Since the objective function of \eqref{optimization_problem} is convex, descent algorithms can be used for its minimization. However, conventional descent methods require exact determination of the gradient of the objective function $\nabla_{\bbw} F(\bbw)=\mbE_{\bbtheta}[\nabla_{\bbw} f(\bbw,{\bbtheta})]$, which is intractable in general. Stochastic gradient descent (SGD) methods overcome this issue by using unbiased gradient estimates based on small subsamples of data and are the workhorse methodology used to solve large-scale stochastic optimization problems \cite{Bottou,Shwartz,Zhang,Roux, Nemirovski}. Practical appeal of SGD remains limited, however, because they need large number of iterations to converge. This problem is most acute when the variable dimension $n$ is large as the condition number tends to increase with $n$. Developing stochastic Newton algorithms, on the other hand, is of little use because unbiased estimates of Newton steps are not easy to compute \cite{Birge}. 

Recourse to quasi-Newton methods then arises as a natural alternative. Indeed, quasi-Newton methods achieve superlinear convergence rates in deterministic settings while relying on gradients to compute curvature estimates \cite{Dennis,Powell,Byrd, Nocedal}. Since unbiased gradient estimates are computable at manageable cost, stochastic generalizations of quasi-Newton methods are not difficult to devise \cite{Schraudolph,AryanAle,Bordes}. Numerical tests of these methods on simple quadratic objectives suggest that stochastic quasi-Newton methods retain the convergence rate advantages of their deterministic counterparts \cite{Schraudolph}. The success of these preliminary experiments notwithstanding, stochastic quasi-Newton methods are prone to yield near singular curvature estimates that may result in erratic behavior (see Section \ref{sec:SVM2}).

In this paper we introduce a stochastic regularized version of the Broyden-Fletcher-Goldfarb-Shanno (BFGS) quasi-Newton method to solve problems with the generic structure in \eqref{optimization_problem}. The proposed regularization avoids the near-singularity problems of more straightforward extensions and yields an algorithm with provable convergence guarantees when the functions $f(\bbw,{\bbtheta})$ are strongly convex. 

We begin the paper with a brief discussion of SGD (Section \ref{sec:problem}) and deterministic BFGS (Section \ref{sec_regularized_bfgs}). The fundamental idea of BFGS is to continuously satisfy a secant condition that captures information on the curvature of the function being minimized while staying close to previous curvature estimates. To regularize deterministic BFGS we retain the secant condition but modify the proximity condition so that eigenvalues of the Hessian approximation matrix stay above a given threshold (Section \ref{sec_regularized_bfgs}). This regularized version is leveraged to introduce the regularized stochastic BFGS algorithm (Section \ref{sec_stochastic_bfgs}). Regularized stochastic BFGS differs from standard BFGS in the use of a regularization to make a bound on the largest eigenvalue of the Hessian inverse approximation matrix and on the use of stochastic gradients in lieu of deterministic gradients for both, the determination of descent directions and the approximation of the objective function's curvature. We abbreviate regularized stochastic BFGS as RES\footnote{The letters ``R and ``E'' appear in ``regularized'' as well as in the names of Broyden, Fletcher, and Daniel Goldfarb; ``S'' is for ``stochastic'' and Shanno.}.

Convergence properties of RES are then analyzed (Section \ref{sec_convergence}). We prove that lower and upper bounds on the Hessians of the sample functions $f(\bbw,{\bbtheta})$ are sufficient to guarantee convergence to the optimal argument $\bbw^*$ with probability 1 over realizations of the sample functions (Theorem \ref{convg}). We complement this result with a characterization of the convergence rate which is shown to be at least linear in expectation (Theorem \ref{theo_convergence_rate}). Linear expected convergence rates are typical of stochastic optimization algorithms and, in that sense, no better than SGD. Advantages of RES relative to SGD are nevertheless significant, as we establish in numerical results for the minimization of a family of quadratic objective functions of varying dimensionality and condition number (Section \ref{sec:simulations}). As we vary the condition number we observe that for well conditioned objectives RES and SGD exhibit comparable performance, whereas for ill conditioned functions RES outperforms SGD by an order of magnitude (Section \ref{sec_numerical_analysis_condition_number}). As we vary problem dimension we observe that SGD becomes unworkable for large dimensional problems. RES however, exhibits manageable degradation  as the number of iterations required for convergence doubles when the problem dimension increases by a factor of ten (Section \ref{sec_numerical_analysis_problem_dimension}).

An important example of a class of problems having the form in \eqref{optimization_problem} are support vector machines (SVMs) that reduce binary classification to the determination of a hyperplane that separates points in a given training set; see, e.g., \cite{Vapnik, Bottou, BGV}.  We adapt RES for SVM  problems (Section \ref{sec:SVMproblem}) and show the improvement relative to SGD in convergence time, stability, and classification accuracy through numerical analysis  (Section\ref{sec:SVM2}). We also compare RES to standard (non-regularized) stochastic BFGS. The regularization in RES is fundamental in guaranteeing convergence as standard (non-regularized) stochastic BFGS is observed to routinely fail in the computation of a separating hyperplane.

\section{Algorithm definition} \label{sec:problem}

Recall the definitions of the sample functions $f(\bbw,\bbtheta)$ and the average function $F(\bbw):=\mbE_{\bbtheta}[f(\bbw,{\bbtheta})]$. We assume the sample functions $f(\bbw,\bbtheta)$ are strongly convex for all $\bbtheta$. This implies the objective function $F(\bbw):=\mbE_{\bbtheta}[f(\bbw,{\bbtheta})]$, being an average of the strongly convex sample functions, is also strongly convex. We can find the optimal argument $\bbw^{*}$ in \eqref{optimization_problem} with a gradient descent algorithm where gradients of $F(\bbw)$ are given by
\begin{align}\label{gradient}
    \bbs(\bbw) := \nabla F(\bbw) 
                  = \mbE_{\bbtheta}[\nabla f(\bbw,{\bbtheta})].
\end{align}
When the number of functions $f(\bbw,\bbtheta)$ is large, as is the case in most problems of practical interest, exact evaluation of the gradient $\bbs(\bbw)$ is impractical. This motivates the use of stochastic gradients in lieu of actual gradients. More precisely, consider a given set of $L$ realizations $\tbtheta=[\bbtheta_{1};...;\bbtheta_{L}]$ and define the stochastic gradient of $F(\bbw)$ at $\bbw$ given samples $\tbtheta$ as
\begin{equation}\label{stochastic_gradient}
   \hbs(\bbw,\tbtheta) 
          := \frac{1}{L}\sum_{l=1}^{L}  \nabla f(\bbw,{\bbtheta_{l}}).
\end{equation}
Introducing now a time index $t$, an initial iterate $\bbw_{0}$, and a step size sequence $\epsilon_{t}$, a stochastic gradient descent algorithm is defined by the iteration
\begin{equation}\label{update_stochastic_gradient}
   \bbw_{t+1} = \bbw_{t}-{\epsilon_{t}} \ \! \hbs(\bbw_t,\tbtheta_t).
\end{equation}
To implement \eqref{update_stochastic_gradient} we compute stochastic gradients $\hbs(\bbw_t,\tbtheta_t)$ using \eqref{stochastic_gradient}. In turn, this requires determination of the gradients of the random functions $ f(\bbw,\bbtheta_{tl})$ for each  $\bbtheta_{tl}$ component of $\tbtheta_t$ and their corresponding average. The computational cost is manageable for small values of $L$. 

The stochastic gradient $\hbs(\bbw,\tbtheta)$ in \eqref{stochastic_gradient} is an unbiased estimate of the (average) gradient $\bbs(\bbw)$ in \eqref{gradient} in the sense that  $\mbE_{\tbtheta}[\hbs(\bbw,\tbtheta)]=\bbs(\bbw)$. Thus, the iteration in \eqref{update_stochastic_gradient} is such that, on average, iterates descend along a negative gradient direction. This intuitive observation can be formalized into a proof of convergence when the step size sequence is selected as nonsummable but square summable, i.e.,
\begin{equation}\label{stepsize_condition}
   {\sum_{t=0}^{\infty} \eps_t = \infty, \quad \text{and} \quad
          \sum_{t=0}^{\infty} \eps_t^2 < \infty }.
\end{equation}
A customary step size choice for which \eqref{stepsize_condition} holds is to make {$\eps_{t}= \eps_{0} T_{0}/(T_{0}+t)$}, for given parameters $\eps_{0}$ and $T_{0}$ that control the initial step size and its speed of decrease, respectively. Convergence notwithstanding, the number of iterations required to approximate $\bbw^*$ is very large in problems that don't have small condition numbers. This motivates the alternative methods we discuss in subsequent sections.
%
%
%
%
%
%
%

%
\subsection{Regularized BFGS}\label{sec_regularized_bfgs}

To speed up convergence of (\ref{update_stochastic_gradient}) resort to second order methods is of little use because evaluating Hessians of the objective function is computationally intensive. A better suited methodology is the use of quasi-Newton methods whereby gradient descent directions are premultiplied by a matrix $\bbB_t^{-1}$,
\begin{equation}\label{eqn_bfgs_descent}
   \bbw_{t+1} = \bbw_{t}-\epsilon_{t}\ \bbB_t^{-1}\bbs(\bbw_t).
\end{equation}
The idea is to select positive definite matrices $\bbB_t\succ0$ close to the Hessian of the objective function $\bbH(\bbw_t):=\nabla^2 F(\bbw_t)$. Various methods are known to select matrices $\bbB_t$, including those by Broyden e.g., \cite{Broyden}; {Davidon, Feletcher, and Powell (DFP) \cite{DFP}; and Broyden, Fletcher, Goldfarb, and Shanno (BFGS) e.g., \cite{Nocedal,Byrd,Powell}. We work here with the matrices $\bbB_t$ used in BFGS since they have been observed to work best in practice \cite{Byrd}.

In BFGS -- and all other quasi-Newton methods for that matter -- the function's curvature is approximated by a finite difference. Specifically, define the variable and gradient variations at time $t$ as
\begin{equation}\label{ball}
      \bbv_{t} := \bbw_{t+1}-\bbw_{t},  \quad \text{and} \quad  
      \bbr_{t}  := \bbs(\bbw_{t+1})-\bbs(\bbw_t),
\end{equation}
respectively, and select the matrix $\bbB_{t+1}$ to be used in the next time step so that it satisfies the secant condition $ \bbB_{t+1} \bbv_{t} = \bbr_{t}$. The rationale for this selection is that {the Hessian $\bbH(\bbw_t)$ satisfies this condition for $\bbw_{t+1}$ tending to $\bbw_{t}$}. Notice however that the secant condition $ \bbB_{t+1} \bbv_{t} = \bbr_{t}$ is not enough to completely specify $\bbB_{t+1}$. To resolve this indeterminacy, matrices $\bbB_{t+1}$ in BFGS are also required to be as close as possible to $\bbB_{t}$ in terms of the Gaussian differential entropy,
\begin{alignat}{2}\label{jaygozin}
   \bbB_{t+1}\ =\ &\argmin_{\bbZ}\ &&\ \tr \left[ \bbB_{t}^{-1}\bbZ \right]-\log\det\left[ \bbB_{t}^{-1}\bbZ\right]-n,\nonumber\\
                  &\st\     &&\  \bbZ\bbv_{t} = \bbr_{t},\quad 
                                 \bbZ \succeq\bbzero.
\end{alignat}
The constraint $\bbZ \succeq\bbzero$ in \eqref{jaygozin} restricts the feasible space to positive semidefinite matrices whereas the constraint $\bbZ \bbv_{t} =  \bbr_{t}$ requires $\bbZ$ to satisfy the secant condition. The objective $\tr(\bbB_{t}^{-1}\bbZ)-\log\det(\bbB_{t}^{-1}\bbZ)-n$ represents the differential entropy between random variables with zero-mean Gaussian distributions $\ccalN(\bbzero,\bbB_{t})$ and $\ccalN(\bbzero,\bbZ)$ having covariance matrices $\bbB_t$ and $\bbZ$. The differential entropy is nonnegative and equal to zero if and only if $\bbZ=\bbB_{t}$. The solution $\bbB_{t+1}$ of the semidefinite program in \eqref{jaygozin} is therefore closest to $\bbB_{t}$ in the sense of minimizing the Gaussian differential entropy among all positive semidefinite matrices that satisfy the secant condition $\bbZ \bbv_{t} =  \bbr_{t}$. 

Strongly convex functions are such that the inner product of the gradient and variable variations is positive, i.e., $\bbv_{t}^T\bbr_{t}>0$. In that case the matrix $\bbB_{t+1}$ in \eqref{jaygozin} is explicitly given by the update -- see, e.g., \cite{Nocedal} and the proof of Lemma \ref{flen} --,
\begin{equation}\label{answer2}
\bbB_{t+1}=\bbB_{t}+ {{\bbr_{t}\bbr_{t}^{T}}\over 
{ \bbv_{t}^{T }\bbr_{t}}} -
{{\bbB_{t} \bbv_{t} \bbv_{t}^{T}\bbB_{t}}\over 
{\bbv_{t}^{T} \bbB_{t}  \bbv_{t}}}.
\end{equation}
In principle, the solution to \eqref{jaygozin} could be positive semidefinite but not positive definite, i.e., we can have $\bbB_{t+1}\succeq\bbzero$ but $\bbB_{t+1}\not\succ\bbzero$. However, through direct operation in \eqref{answer2} it is not difficult to conclude that $\bbB_{t+1}$ stays positive definite if the matrix $\bbB_t$ is positive definite. Thus, initializing the curvature estimate with a positive definite matrix $\bbB_{0}\succ\bbzero$ guarantees $\bbB_{t}\succ\bbzero$ for all subsequent times $t$. Still, it is possible for the smallest eigenvalue of $\bbB_t$ to become arbitrarily close to zero which means that the largest eigenvalue of $\bbB_{t}^{-1}$ can become arbitrarily large. This has been proven not to be an issue in BFGS implementations but is a more significant challenge in the stochastic version proposed here. 

To avoid this problem we introduce a regularization of \eqref{jaygozin} to enforce the eigenvalues of $\bbB_{t+1}$ to exceed a positive constant $\delta$. Specifically, we redefine $\bbB_{t+1}$ as the solution of the semidefinite program,
\begin{alignat}{2}\label{jadid}
   \bbB_{t+1} \!= 
      &\argmin_{\bbZ}\ &&\ \! \tr \!\left[ \bbB_{t}^{-1}(\bbZ\!\!-\!\delta \bbI) \right]
                             \!-\!\log\det \! \left[ \bbB_{t}^{-1}(\bbZ\!\!-\!\delta \bbI) \right] \!-\!n,\nonumber\\
      &\st      && \bbZ \bbv_{t} = \bbr_{t},\quad 
                                     \bbZ \succeq\bbzero.
\end{alignat}
The curvature approximation matrix $\bbB_{t+1}$ defined in \eqref{jadid} still satisfies the secant condition $\bbB_{t+1} \bbv_{t} = \bbr_{t}$ but has a different proximity requirement since instead of comparing $\bbB_t$ and $\bbZ$ we compare $\bbB_t$ and $\bbZ-\delta \bbI$. While \eqref{jadid} does not ensure that all eigenvalues of $\bbB_{t+1}$ exceed $\delta$ we can show that this will be the case under two minimally restrictive assumptions. We do so in the following proposition where we also give an explicit solution for \eqref{jadid} analogous to the expression in \eqref{answer2} that solves the non regularized problem in \eqref{jaygozin}.

%
\begin{proposition}\label{flen}
Consider the semidefinite program in \eqref{jadid} where the matrix $\bbB_{t}\succ\bbzero$ is positive definite and define the corrected gradient variation 
\begin{equation}\label{corrected_gradient_variation}
\tbr_t:=\bbr_t-\delta\bbv_t.
\end{equation}
If the inner product $\tbr_{t}^{T}\bbv_{t}=(\bbr_t-\delta\bbv_t)^T\bbv_t>0$ is positive, the solution $\bbB_{t+1}$ of \eqref{jadid} is such that all eigenvalues of $\bbB_{t+1}$ are larger than $\delta$,
\begin{equation}\label{min_eigenvalue_condition}
 \bbB_{t+1} \succeq \delta \bbI.
\end{equation} 
Furthermore, $\bbB_{t+1}$ is explicitly given by the expression
\begin{equation}\label{eqn_inverse_appx_update}
\bbB_{t+1}=\bbB_{t} + {{ \tbr_t  \tbr_t^{T}}\over{\bbv_{t}^{T} \tbr_t}}- {{\bbB_{t} \bbv_{t}\bbv_{t}^{T}{\bbB_{t}} }\over{\bbv_{t}^{T}\bbB_{t}\bbv_{t}}} +\delta \bbI.
\end{equation} \end{proposition}

%
\begin{myproof} See Appendix. \end{myproof}

%
Comparing \eqref{answer2} and \eqref{eqn_inverse_appx_update} it follows that the differences between BFGS and regularized BFGS are the replacement of the gradient variation $\bbr_t$ in \eqref{ball} by the corrected variation $\tbr_t:=\bbr_t-\delta\bbv_t$ and the addition of the regularization term $\delta\bbI$. We use \eqref{eqn_inverse_appx_update} in the construction of the stochastic BFGS algorithm in the following section.


%
\subsection{RES: Regularized Stochastic BFGS}\label{sec_stochastic_bfgs}

{As can be seen from \eqref{eqn_inverse_appx_update} the regularized BFGS curvature estimate $\bbB_{t+1}$ is obtained as a function of previous estimates $\bbB_t$, iterates $\bbw_t$ and $\bbw_{t+1}$, and corresponding gradients $\bbs(\bbw_t)$ and $\bbs(\bbw_{t+1})$. We can then think of a method in which gradients $\bbs(\bbw_t)$ are replaced by stochastic gradients $\hbs(\bbw_t,\tbtheta_t)$ in both, the curvature approximation update in \eqref{eqn_inverse_appx_update} and the descent iteration in \eqref{eqn_bfgs_descent}.}{ Specifically, start at time $t$ with current iterate $\bbw_t$ and let $\hbB_{t}$ stand for the Hessian approximation computed by stochastic BFGS in the previous iteration. Obtain a batch of samples $\tbtheta_t=[\bbtheta_{t1};...;\bbtheta_{tL}]$, determine the value of the stochastic gradient $\hbs(\bbw_{t},\tbtheta_{t})$ as per \eqref{stochastic_gradient}, and update the iterate $\bbw_{t}$ as }
\begin{equation}\label{eqn_sbfgs_dual_iteration}
   \bbw_{t+1} 
       = \bbw_{t} 
            - \epsilon_{t} \left(\hbB_{t}^{-1}+\Gamma \bbI\right) \hbs(\bbw_{t},\tbtheta_{t}),
\end{equation}
where we added the identity bias term $\Gamma \bbI$ for a given positive constant $\Gamma>0$.
{Relative to SGD as defined by \eqref{update_stochastic_gradient}, RES as defined by \eqref{eqn_sbfgs_dual_iteration} differs in the use of the matrix $\hbB_{t}^{-1}+\Gamma \bbI$ to account for the curvature of $F(\bbw)$. Relative to (regularized or non regularized) BFGS as defined in \eqref{eqn_bfgs_descent} RES differs in the use of stochastic gradients $\hbs(\bbw_{t},\tbtheta_{t})$ instead of actual gradients and in the use of the curvature approximation $\hbB_{t}^{-1}+\Gamma \bbI$ in lieu of $\bbB_t^{-1}$.} {Observe that in \eqref{eqn_sbfgs_dual_iteration} we add a bias $\Gamma\bbI$ to the curvature approximation $\hbB_{t}^{-1}$. This is necessary to ensure convergence by hedging against random variations in $\hbB_{t}^{-1}$ as we discuss in Section \ref{sec_convergence}.} 

To update the Hessian approximation matrix $\hbB_t$ compute the stochastic gradient $\hbs(\bbw_{t+1},\tbtheta_{t})$ associated with the {\it same} set of samples $\tbtheta_t$ used to compute the stochastic gradient $\hbs(\bbw_{t},\tbtheta_{t})$. Define then the stochastic gradient variation at time $t$ as
\begin{equation}\label{chris}
   \hbr_{t} :=\hbs(\bbw_{t+1},\tbtheta_{t})-\hbs(\bbw_{t},\tbtheta_{t}),
\end{equation}
and redefine $\tbr_{t}$ so that it stands for the modified stochastic gradient variation
\begin{equation}\label{chris2}
   \tbr_{t} := \hbr_{t} - \delta \bbv_{t},
\end{equation}
by using $\hbr_{t}$ instead of $\bbr_{t}$.
The Hessian approximation $\hbB_{t+1}$ for the next iteration is defined as the matrix that satisfies the stochastic secant condition $\bbZ \bbv_{t} =  \hbr_{t}$ and is closest to $\hbB_t$ in the sense of \eqref{jadid}. As per Proposition \ref{flen} we can compute $\hbB_{t+1}$ explicitly as 
\begin{equation}\label{akbar}
   \hbB_{t+1} = \hbB_{t} + {{\tbr_{t}\tbr_{t}^{T}}\over{\bbv_{t}^{T}\tbr_{t}}}
- {{\hbB_{t} \bbv_{t}\bbv_{t}^{T}{\hbB_{t}} }\over{\bbv_{t}^{T}\hbB_{t}\bbv_{t}}} +\delta \bbI .
\end{equation}
as long as $(\hbr_t-\delta\bbv_t)^T\bbv_t = \tbr^T\bbv_t>0$. Conditions to guarantee that $\tbr_{t}^T\bbv_t>0$ are introduced in Section \ref{sec_convergence}. 

%
{\small \begin{algorithm}[t] 
\caption{RES: Regularized Stochastic BFGS}
{\small \label{algo_stochastic_bfgs} 
\begin{algorithmic}[1]
\Require Variable $\bbw_0$. Hessian approximation  $\hbB_0 \succ\delta \bbI$.
\For {$t=0,1,2,\ldots$}
   \State Acquire $L$ independent samples $\tbtheta_t=[\bbtheta_{t1},\ldots,\bbtheta_{tL}]$
   \State Compute $\hbs(\bbw_{t},\tbtheta_{t})$ [cf. \eqref{stochastic_gradient}]
\begin{equation*}
   \hbs(\bbw_{t},\tbtheta_{t}) 
          = \frac{1}{L}\sum_{l=1}^{L}  \nabla_{\bbw} f(\bbw_{t},{\bbtheta_{tl}}).
\end{equation*}

   \State Descend along direction $(\hbB_{t}^{-1}+\Gamma \bbI)\ \! \hbs(\bbw_{t},\tbtheta_{t})$ 
          [cf. \eqref{eqn_sbfgs_dual_iteration}]
          \begin{equation*}
             \bbw_{t+1}=\bbw_{t}-\epsilon_{t}\ (\hbB_{t}^{-1}+\Gamma \bbI)\ \! \hbs(\bbw_{t},\tbtheta_{t}).
          \end{equation*}
   \State Compute $\hbs(\bbw_{t+1},\tbtheta_{t})$ [cf. \eqref{stochastic_gradient}]
\begin{equation*}
   \hbs(\bbw_{t+1},\tbtheta_{t}) 
          = \frac{1}{L}\sum_{l=1}^{L}  \nabla_{\bbw} f(\bbw_{t+1},{\bbtheta_{tl}}).
\end{equation*}
   \State Compute variable variation [cf. \eqref{ball}]
          \begin{equation*}
             \bbv_{t} = \bbw_{t+1}-\bbw_{t}. 
          \end{equation*}
          \State Compute modified stochastic gradient variation [cf. \eqref{chris2}]
          \begin{equation*}
             \tbr_{t}=\hbs(\bbw_{t+1},\tbtheta_{t})-\hbs(\bbw_{t},\tbtheta_{t})-\delta \bbv_{t}.
          \end{equation*}
   \State Update Hessian approximation matrix
          [cf. \eqref{akbar}]
          \begin{equation*}            
  					 \hbB_{t+1} = \hbB_{t} + 			  		{{\tbr_{t}\tbr_{t}^{T}}\over{\bbv_{t}^{T}\tbr_{t}}}
	- {{\hbB_{t} \bbv_{t}\bbv_{t}^{T}{\hbB_{t}} 	}\over{\bbv_{t}^{T}\hbB_{t}\bbv_{t}}} +\delta \bbI .
          \end{equation*}

\EndFor
\end{algorithmic}}
\end{algorithm}}


The resulting RES algorithm is summarized in Algorithm \ref{algo_stochastic_bfgs}. The two core steps in each iteration are the descent in Step 4 and the update of the Hessian approximation $\hbB_{t}$ in Step 8. Step 2 comprises the observation of $L$ samples that are required to compute the stochastic gradients in steps 3 and 5. The stochastic gradient $\hbs(\bbw_{t},\tbtheta_{t})$ in Step 3 is used in the descent iteration in Step 4. The stochastic gradient of Step 3 along with the stochastic gradient $\hbs(\bbw_{t+1},\tbtheta_{t})$ of Step 5 are used to compute the variations in steps 6 and 7 that permit carrying out the update of the Hessian approximation $\hbB_{t}$ in Step 8. Iterations are initialized at arbitrary variable $\bbw_0$ and positive definite  matrix $\hbB_{0}$ with the smallest eigenvalue larger than $\delta$.

%
\begin{remark}\normalfont One may think that the natural substitution of the gradient variation $\bbr_{t} = \bbs(\bbw_{t+1})-\bbs(\bbw_t)$ is the stochastic gradient variation $\hbs(\bbw_{t+1},\tbtheta_{t+1})-\hbs(\bbw_{t},\tbtheta_{t})$ instead of the variation $\hbr_{t}=\hbs(\bbw_{t+1},\tbtheta_{t})-\hbs(\bbw_{t},\tbtheta_{t})$ in \eqref{chris}. This would have the advantage that $\hbs(\bbw_{t+1},\tbtheta_{t+1})$ is the stochastic gradient used to descend in iteration $t+1$ whereas $\hbs(\bbw_{t+1},\tbtheta_{t})$ is not and is just computed for the purposes of updating $\bbB_t$. Therefore, using the variation $\hbr_{t}=\hbs(\bbw_{t+1},\tbtheta_{t})-\hbs(\bbw_{t},\tbtheta_{t})$ requires twice as many stochastic gradient evaluations as using the variation  $\hbs(\bbw_{t+1},\tbtheta_{t+1})-\hbs(\bbw_{t},\tbtheta_{t})$. However, the use of the variation $\hbr_{t}=\hbs(\bbw_{t+1},\tbtheta_{t})-\hbs(\bbw_{t},\tbtheta_{t})$ is necessary to ensure that $(\hbr_t-\delta\bbv_t)^T\bbv_t = \tbr_{t}^T\bbv_t>0$, which in turn is required for \eqref{akbar} to be true. This cannot be guaranteed if we use the variation $\hbs(\bbw_{t+1},\tbtheta_{t+1})-\hbs(\bbw_{t}\tbtheta_{t})$ -- see Lemma \ref{lecce} for details. The same observation holds true for the non-regularized version of stochastic BFGS introduced in \cite{Schraudolph}.
\end{remark}

\section{Convergence}\label{sec_convergence}

For the subsequent analysis it is convenient to define the instantaneous objective function associated with samples $\tbtheta=[\bbtheta_1,\ldots,\bbtheta_L]$ as
\begin{equation}\label{eqn_isntantaneous_dual_function}
   \hhatf(\bbw,{\tbtheta}) := \frac{1}{L}\sum_{l=1}^L f(\bbw,{\bbtheta_{l}}).
\end{equation}
The definition of the instantaneous objective function $ \hhatf(\bbw,{\tbtheta})$ in association with  the fact that $F(\bbw) := \mbE_{\bbtheta}[f(\bbw,{\bbtheta})]$ implies
\begin{equation}\label{eqn_dual_function_as_expectation_of_isntantaneous_dual_function}
   F(\bbw) = \mbE_{\bbtheta}[\hhatf(\bbw,{\tbtheta})].
\end{equation}
Our goal here is to show that as time progresses the sequence of variable iterates $\bbw_t$ approaches the optimal argument $\bbw^*$. In proving this result we make the following assumptions.

%
\begin{assumption}\label{ass_intantaneous_hessian_bounds}\normalfont  The instantaneous functions $\hhatf(\bbw,{\tbtheta})$ are twice differentiable and the eigenvalues of the instantaneous Hessian $\hat{ \bbH}(\bbw,{\tbtheta} )=\nabla_{\bbw}^2\hhatf(\bbw,{\tbtheta})$ are  bounded between constants $0<\tdm$ and $\tdM<\infty$ for all random variables $\tbtheta$,
\begin{equation}\label{hassan}
   \tdm\bbI \ \preceq\ \hat{ \bbH}(\bbw,{\tbtheta}) \ \preceq \ \tdM \bbI.
\end{equation} \end{assumption}

%
\begin{assumption}\normalfont\label{ass_bounded_stochastic_gradient_norm} The second moment of the norm of the stochastic gradient is bounded for all $\bbw$. i.e., there exists a constant $S^2$ such that for all variables $\bbw$ it holds
\begin{equation}\label{ekhtelaf}
   \mbE_{\bbtheta} \big{[} \| \hbs(\bbw_{t},\tbtheta_{t})\|^{2} \big{]} \leq S^{2}, 
\end{equation} \end{assumption}

%
\begin{assumption}\normalfont\label{ass_delta_less_m} The regularization constant $\delta$ is smaller than the smallest Hessian eigenvalue $\tdm$, i.e., $\delta<\tdm$.
\end{assumption}

%
As a consequence of Assumption \ref{ass_intantaneous_hessian_bounds} similar eigenvalue bounds hold for the (average) function $F(\bbw)$. Indeed, it follows from the linearity of the expectation operator and the expression in \eqref{eqn_dual_function_as_expectation_of_isntantaneous_dual_function} that the Hessian is $\nabla_{\bbw}^{2}F(\bbw)= \bbH(\bbw)=\mbE_{\bbtheta}[\hat{\bbH}(\bbw,{\tbtheta})]$. Combining this observation with the bounds in \eqref{hassan} it follows that there are constants $m\geq\tdm$ and $M\leq\tdM$ such that
\begin{equation}\label{bbb}
   \tdm\bbI \ \preceq\ m\bbI\ \preceq\ \bbH(\bbw) \ \preceq  M\bbI \ \preceq \ \tdM\bbI.
\end{equation} 
The bounds in \eqref{bbb} are customary in convergence proofs of descent methods. For the results here the stronger condition spelled in Assumption \ref{ass_intantaneous_hessian_bounds} is needed. The restriction imposed by Assumption \ref{ass_bounded_stochastic_gradient_norm} is typical of stochastic descent algorithms, its intent being to limit the random variation of stochastic gradients. Assumption \ref{ass_delta_less_m} is necessary to guarantee that the inner product $\tbr_{t}^{T}\bbv_{t}=(\bbr_t-\delta\bbv_t)^T\bbv_t>0$ [cf. Proposition \ref{flen}] is positive as we show in the following lemma.

%
\begin{lemma}\label{lecce}
Consider the modified stochastic gradient variation $\tbr_{t}$ defined in \eqref{chris2} and the variable variation $\bbv_{t}$ defined in \eqref{ball}. Let Assumption \ref{ass_intantaneous_hessian_bounds} hold and recall the lower bound $\tdm$ on the smallest eigenvalue of the instantaneous Hessians. Then, for all constants $\delta < \tdm$ it holds
\begin{equation}\label{claim233}
   \tbr_{t}^{T}\bbv_{t} \ =\ (\hbr_{t}-\delta \bbv_{t})^T\bbv_{t}\ \geq (\tdm-\delta) \|\bbv_{t}\|^{2} \ >\ 0.
\end{equation}\end{lemma}

%
\begin{myproof} 
As per \eqref{hassan} in Assumption 1 the eigenvalues of the instantaneous Hessian $\hat{ \bbH}(\bbw,{\tbtheta})$ are bounded by $\tdm$ and $\tdM$. Thus, for any given vector $\bbz$ it holds
\begin{equation}\label{liverpool}
\tdm \|\bbz\|^{2} \leq \bbz^{T} \hat{ \bbH}(\bbw,{\tbtheta}) \bbz \leq \tdM \|\bbz\|^{2}.
\end{equation}
For given $\bbw_{t}$ and $\bbw_{t+1}$ define the mean instantaneous Hessian $\hbG_{t}$ as the average Hessian value along the segment $[\bbw_{t},\bbw_{t+1}]$
\begin{equation}\label{average}
\hbG_{t}= \int_{0}^{1}  \hat{ \bbH} \left( \bbw_{t}+\tau(\bbw_{t+1}-\bbw_{t}),{\tbtheta_{t}}\right) d\tau.
\end{equation}
Consider now the instantaneous gradient $\hbs( \bbw_{t}+\tau(\bbw_{t+1}-\bbw_{t}),\tbtheta_{t})$ evaluated at $\bbw_{t}+\tau(\bbw_{t+1}-\bbw_{t})$ and observe that its derivative with respect to $\tau$ is $\partial\hbs\big( \bbw_{t}+\tau(\bbw_{t+1}-\bbw_{t}),\tbtheta_{t}\big)/\partial \tau = \hbH(\bbw_{t}+\tau( \bbw_{t+1} - \bbw_{t}),{\tbtheta_{t}})  (\bbw_{t+1} - \bbw_{t})$. It then follows from the fundamental theorem of calculus that 
\begin{align}\label{lazio}
   \int_{0}^{1}\!\hbH ( \bbw_{t}+\tau(\bbw_{t+1}-\bbw_{t}),{\tbtheta_{t}} )  
                        (\bbw_{t+1} - \bbw_{t})\ d\tau  
      = \nonumber \\
       \hbs(\bbw_{t+1},\tbtheta_{t}) - \hbs(\bbw_{t},\tbtheta_{t}) .
\end{align}
Using the definitions of the mean instantaneous Hessian $\hbG_{t}$ in \eqref{average} as well as the definitions of the stochastic gradient variations $\hbr_{t}$ and variable variations $\bbv_{t}$ in \eqref{chris} and \eqref{ball} we can rewrite \eqref{lazio} as
\begin{equation}\label{mancity}
 \hbG_{t} \bbv_{t} = \hbr_{t}.
\end{equation}
Invoking \eqref{liverpool} for the integrand in \eqref{average}, i.e., for $\hat{ \bbH}(\bbw,{\tbtheta}) = \hat{ \bbH}\big{(}\bbw_{t}+\tau(\bbw_{t+1}-\bbw_{t}),{\tbtheta} \big{)}$, it follows that for all vectors $\bbz$ the mean instantaneous Hessian $\hbG_{t}$ satisfies
\begin{equation}\label{arsenal}
   \tdm \|\bbz\|^{2} \leq \bbz^{T} \hbG_{t} \bbz \leq \tdM \|\bbz\|^{2}.
\end{equation}
{The claim in \eqref{claim233} follows from \eqref{mancity} and \eqref{arsenal}.} Indeed, consider the ratio of inner products $\hbr_{t}^{T}\bbv_{t}/\bbv_{t}^{T}\bbv_{t}$ and use \eqref{mancity} and the first inequality in \eqref{arsenal} to write
\begin{equation}\label{jose}
   \frac{\hbr_{t}^{T}\bbv_{t}}{\bbv_{t}^{T}\bbv_{t}}
       = \frac{{\bbv_{t}^{T} \hbG_{t} \bbv_{t} }}{\bbv_{t}^{T} \bbv_{t}} \geq \tdm.
\end{equation}
%
Consider now the inner product $\tbr_{t}^{T}\bbv_{t} \ =\ (\hbr_{t}-\delta \bbv_{t})^T\bbv_{t}$ in \eqref{claim233} and use the bound in \eqref{jose} to write
\begin{align}\label{bound_for_norm_r}
   \tbr_{t}^{T}\bbv_{t}
      \ =   \  \hbr_{t}^{T}\bbv_{t} - \delta  \bbv_{t}^{T}\bbv_{t}
      \ \geq\  {{\tdm}}\bbv_{t}^{T}\bbv_{t} - \delta  \bbv_{t}^{T}\bbv_{t}
\end{align}
Since we are selecting $\delta < \tdm$ by hypothesis it follows that \eqref{claim233} is true for all times $t$. \end{myproof}

%
Initializing the curvature approximation matrix $\hbB_0\succ\delta\bbI$, which implies $\hbB_0^{-1}\succ\bbzero$, and setting $\delta < \tdm$ it follows from Lemma \ref{lecce} that the hypotheses of Proposition \ref{flen} are satisfied for $t=0$. Hence, the matrix $\hbB_1$ computed from \eqref{akbar} is the solution of the semidefinite program in \eqref{jadid} and, more to the point, satisfies $\hbB_1\succ\delta\bbI$, which in turn implies  $\hbB_1^{-1}\succ\bbzero$. Proceeding recursively we can conclude that $\hbB_{t}\succ\delta\bbI\succ\bbzero$ for all times $t\geq0$. Equivalently, this implies that all the eigenvalues of $\hbB_t^{-1}$ are between $0$ and $1/\delta$ and that, as a consequence, the matrix $\hbB_t^{-1}+\Gamma \bbI$ is such that
\begin{equation}\label{eqn_eigenvalue_critical_bounds}
   \Gamma \bbI
      \ \preceq\ \hbB_t^{-1}+\Gamma \bbI  
      \ \preceq\ \left(\Gamma+\frac{1}{\delta}\right) \bbI.
\end{equation} 
Having matrices $\hbB_t^{-1}+\Gamma \bbI$ that are strictly positive definite with eigenvalues uniformly upper bounded by $\Gamma+(1/\delta)$ leads to the conclusion that if $\hbs(\bbw_{t},\tbtheta_{t})$ is a descent direction, the same holds true of $(\hbB_t^{-1}+\Gamma \bbI)\ \! \hbs(\bbw_{t},\tbtheta_{t})$. The stochastic gradient $\hbs(\bbw_{t},\tbtheta_{t})$ is not a descent direction in general, but we know that this is true for its conditional expectation $\mbE[\hbs(\bbw_{t},\tbtheta_{t}) \given \bbw_{t}] = \nabla_{\bbw} F(\bbw_{t})$. Therefore, we conclude that $(\hbB_t^{-1}+\Gamma \bbI)\ \! \hbs(\bbw_{t},\tbtheta_{t})$ is an average descent direction because $\mbE[(\hbB_t^{-1}+\Gamma \bbI)\ \!\hbs(\bbw_{t},\tbtheta_{t}) \given \bbw_{t}] = (\hbB_t^{-1}+\Gamma \bbI)\ \! \nabla_{\bbw} F(\bbw_{t})$. Stochastic optimization algorithms whose displacements $\bbw_{t+1}-\bbw_t$ are descent directions on average are expected to approach optimal arguments in a sense that we specify formally in the following lemma.

%
\begin{lemma}\label{helpful}
Consider the RES algorithm as defined by \eqref{eqn_sbfgs_dual_iteration}-\eqref{akbar}. If assumptions \ref{ass_intantaneous_hessian_bounds}, \ref{ass_bounded_stochastic_gradient_norm} and \ref{ass_delta_less_m} hold true, the sequence of average function $F(\bbw_{t})$ satisfies
\begin{equation}\label{pedarsag}
    \E{F(\bbw_{t+1})\given \bbw_{t}}  
         \leq F(\bbw_{t})
          -  \epsilon_{t} \Gamma \| \nabla F(\bbw_{t})\|^{2} 
          + K \epsilon_{t}^{2}
\end{equation}
where the constant $K:={MS^{2}} ({1/\delta}+\Gamma)^{2} /2$.

\end{lemma}

%
\begin{myproof} 
As it follows from Assumption 1 the eigenvalues of the Hessian $\bbH(\bbw_{t})=\mbE_{\tbtheta}[\hbH(\bbw_{t},\tbtheta_{t})] = \nabla_{\bbw}^{2} F(\bbw_{t})$ are bounded between $0<m$ and $M<\infty$ as stated in \eqref{bbb}. Taking a Taylor's expansion of the dual function $F(\bbw)$ around $\bbw=\bbw_{t}$ and using the upper bound in the Hessian eigenvalues we can write
\begin{equation}\label{taylor_upper_bound}
   F(\bbw_{t+1}) \leq F(\bbw_{t}) +\nabla F(\bbw_{t})^{T}\!(\bbw_{t+1}-\bbw_{t})   
   + {{M}\over{2}}\|{\bbw_{t+1}-\bbw_{t}}\|^{2}
\end{equation}
From the definition of the RES update in \eqref{eqn_sbfgs_dual_iteration} we can write the difference of two consecutive variables   $\bbw_{t+1}-\bbw_{t} $ as $ -\epsilon_{t}(\hbB_{t}^{-1}+\Gamma \bbI )\ \hbs(\bbw_{t},\tbtheta_{t})$. Making this substitution in \eqref{taylor_upper_bound}, taking expectation with $\bbw_{t}$ given in both sides of the resulting inequality, and observing the fact that when $\bbw_{t}$ is given the Hessian approximation $\hbB_{t}^{-1}$ is deterministic we can write
\begin{align}\label{pedar_sag}
  &  \E{F(\bbw_{t+1})\given \bbw_{t}}  \leq\  F(\bbw_{t})  \\ & \qquad\qquad\qquad 
          - \epsilon_{t}\nabla F(\bbw_{t})^{T}(\hbB_{t}^{-1}+\Gamma \bbI )
                 \E{\hbs(\bbw_{t},\tbtheta_{t})\given\bbw_{t}} \nonumber\\&\qquad\qquad\qquad
          + \frac{\epsilon^{2}M}{2} \ \!
              \E{\left\|(\hbB_{t}^{-1}+\Gamma \bbI )\hbs(\bbw_{t},\tbtheta_{t})\right\|^{2} 
                    \given \bbw_{t}}. \nonumber
\end{align}
We proceed to bound the third term in the right hand side of \eqref{pedar_sag}. Start by observing that the 2-norm of a product is not larger than the product of the 2-norms and that, as noted above, with $\bbw_{t}$ given the matrix $\hbB_{t}^{-1}$ is also given to write 
\begin{align}\label{pedar_salavati}
    &\E{\left\| \left(\hbB_{t}^{-1}+\Gamma \bbI\right)\hbs(\bbw_{t},\tbtheta_{t})\right\|^{2} 
            \given \bbw_{t}}  \nonumber\\  &\hspace{24mm} 
     \leq  \left\|\hbB_{t}^{-1}+\Gamma \bbI \right\|^{2}   
           \E{\left\|  \hbs(\bbw_{t},\tbtheta_{t})\right\|^{2} \given \bbw_{t}}.
\end{align}
Notice that, as stated in \eqref{eqn_eigenvalue_critical_bounds}, $\Gamma + 1/\delta$ is an upper bound for the eigenvalues of $\hbB_{t}^{-1}+\Gamma \bbI$. Further observe that the second moment of the norm of the stochastic gradient is bounded by $  \E{\| \hbs(\bbw_{t},\tbtheta_{t})\|^{2} \given \bbw_{t}} \leq S^{2}$, as stated in Assumption 2. These two upper bounds  substituted in \eqref{pedar_salavati} yield
\begin{equation}\label{dash_akol}
    \E{\left\| \left(\hbB_{t}^{-1}+\Gamma \bbI\right)\hbs(\bbw_{t},\tbtheta_{t})\right\|^{2} 
           \given \bbw_{t}}\leq S^2(1/\delta+\Gamma)^{2}.
\end{equation}
Substituting the upper bound in \eqref{dash_akol} for the third term of \eqref{pedar_sag} and further using the fact that $ \E{\hbs(\bbw_{t},\tbtheta_{t})\given\bbw_{t}}=\nabla F(\bbw_{t})$ in the second term leads to
\begin{align}\label{lavashak}
    \E{F(\bbw_{t+1})\given \bbw_{t}}  
          \leq & F(\bbw_{t})   -  \epsilon_{t} \nabla F(\bbw_{t})^{T}\! \left(\hbB_{t}^{-1}+\Gamma \bbI\right)       \! \nabla F(\bbw_{t}) 
        \nonumber \\ &
          \quad  
          +\frac{\epsilon_{t}^{2}MS^{2}}{{2}} (1/\delta+\Gamma)^{2}  .
\end{align}
We now find a lower bound for the second term in the right hand side of \eqref{lavashak}. Since the Hessian approximation matrices $\hbB_{t}$ are positive definite their inverses $\hbB_{t}^{-1}$ are positive semidefinite. In turn, this implies that all the eigenvalues of $\hbB_{t}^{-1}+ \Gamma \bbI$ are not smaller than $\Gamma$ since $\Gamma \bbI$ increases all the eigenvalues of $\hbB_{t}^{-1}$ by $\Gamma$. This lower bound for the eigenvalues of $\hbB_{t}^{-1}+ \Gamma \bbI$ implies that
\begin{equation}\label{inner_product_lower_bound}
   \nabla F(\bbw_{t})^{T} \left(\hbB_{t}^{-1}+\Gamma \bbI\right)\nabla F(\bbw_{t}) 
       \geq \Gamma \|\nabla F(\bbw_{t}) \|^{2} 
\end{equation}  
Substituting the lower bound in \eqref{inner_product_lower_bound} for the corresponding summand in \eqref{lavashak} and further noting the definition of $K:={MS^{2}} ({1/\delta}+\Gamma)^{2} /2$  in the statement of the lemma, the result in \eqref{taylor_upper_bound} follows.

\end{myproof}

Setting aside the term $K\eps_t^2$ for the sake of argument \eqref{pedarsag} defines a supermartingale relationship for the sequence of average functions $F(\bbw_{t})$. This implies that the sequence $\eps_t\Gamma\ \| \nabla F(\bbw_{t})\|^{2}$ is almost surely summable which, given that the step sizes $\eps_t$ are nonsummable as per \eqref{stepsize_condition}, further implies that the limit infimum $\liminf_{t\to\infty}\|\nabla F(\bbw_{t})\|$ of the gradient norm $\|\nabla F(\bbw_{t})\|$ is almost surely null. This latter observation is equivalent to having $\liminf_{t\to\infty}\| \bbw_{t}-\bbw^{*} \|^{2} =0$ with probability 1 over realizations of the random samples $\{\tbtheta_t\}_{t=0}^\infty$. The term $K\eps_t^2$ is a relatively minor nuisance that can be taken care with a technical argument that we present in the proof of the following theorem.

%
\begin{thm}\label{convg}
Consider the RES algorithm as defined by \eqref{eqn_sbfgs_dual_iteration}-\eqref{akbar}. If assumptions \ref{ass_intantaneous_hessian_bounds}, \ref{ass_bounded_stochastic_gradient_norm} and \ref{ass_delta_less_m} hold true and the sequence of stepsizes satisfies \eqref{stepsize_condition}, the limit infimum of the squared Euclidean distance to optimality $\| \bbw_{t}-\bbw^{*} \|^{2}$ satisfies
\begin{equation}\label{eqn_convg}
   \liminf_{t \to \infty }\| \bbw_{t}-\bbw^{*} \|^{2} = 0 \qquad \text{a.s.}
\end{equation} 
over realizations of the random samples $\{\tbtheta_t\}_{t=0}^\infty$.\end{thm}

%
\begin{myproof} 
The proof uses the relationship in the statement \eqref{pedarsag} of Lemma \ref{helpful} to build a supermartingale sequence. For that purpose define the stochastic process $\gamma_t$ with values 
\begin{equation}\label{alp}
   \gamma_t := F(\bbw_{t})+ K \sum_{u=t}^{\infty}  {{\epsilon_{u}^{2}}}.
\end{equation}
Observe that $\gamma_t$ is well defined because the $\sum_{u=t}^{\infty}  {{\epsilon_{u}^{2}}}<\sum_{u=0}^{\infty}  {{\epsilon_{u}^{2}}}<\infty$ is summable. Further define the sequence $\beta_t$ with values
\begin{equation}\label{beta}
   \beta_t :=\ \epsilon_{t}\ \Gamma\ \| \nabla F(\bbw_{t})\|^{2} .
\end{equation}
Let now $\ccalF_t$ be a sigma-algebra measuring $\gamma_t$, $\beta_t$, and $\bbw_t$. The conditional expectation of $\gamma_{t+1}$ given $\ccalF_t$ can be written as
\begin{equation}\label{eqn_theo_convg_pf_30}
   \E{\gamma_{t+1} \given \ccalF_t} 
       =  \E{F(\bbw_{t}) \given \ccalF_t}
             + K \sum_{u=t}^{\infty}  {{\epsilon_{u}^{2}}},
\end{equation}
because the term $K \sum_{u=t}^{\infty}  {{\epsilon_{u}^{2}}}$ is just a deterministic constant. Substituting \eqref{pedarsag} of Lemma \ref{helpful} into \eqref{eqn_theo_convg_pf_30} and using the definitions of $\gamma_t$ in \eqref{alp} and $\beta_t$ in \eqref{beta} yields
\begin{equation}\label{martin}
   \E{\gamma_{t+1} \given \gamma_t} \ \leq\ \gamma_t - \beta_t
\end{equation}
Since the sequences $\gamma_t$ and $\beta_t$ are nonnegative it follows from \eqref{martin} that they satisfy the conditions of the supermartingale convergence theorem -- {see e.g. theorem E$7.4$ \cite{Solo}} . Therefore, we conclude that: (i) The sequence $\gamma_t$ converges almost surely. (ii) The sum $\sum_{t=0}^{\infty}\beta_t < \infty$ is almost surely finite. Using the explicit form of $\beta_t$ in \eqref{beta} we have that $\sum_{t=0}^{\infty}\beta_t < \infty$ is equivalent to
\begin{equation}\label{psg}
   \sum_{t=0}^{\infty} \epsilon_{t}\Gamma\| \nabla F(\bbw_{t})\|^{2} < \infty, 
       \qquad\text{a.s.}
\end{equation}
Since the sequence of stepsizes is nonsummable for \eqref{psg} to be true we need to have a vanishing subsequence embedded in $\| \nabla F(\bbw_{t})\|^{2}$. By definition, this miles that the limit infimum of the sequence $\| \nabla F(\bbw_{t})\|^{2}$ is null,
\begin{equation}\label{pppp}
   \liminf_{t \to \infty}   \| \nabla F(\bbw_{t})\|^{2}  =  0,  \qquad\text{a.s.}
\end{equation}
To transform the gradient bound in \eqref{pppp} into a bound pertaining to the squared distance to optimality  $\| \bbw_{t} -\bbw^{*} \|^2$ simply observe that the lower bound $m$ on the eigenvalues of $\textbf{H}(\bbw_{t})$ applied to a Taylor's expansion around the optimal argument $\bbw^*$ implies that
\begin{equation}\label{lower_bound_taylor_expansion}
F(\bbw^*)\ \geq\ F(\bbw_{t})+  \nabla F(\bbw_{t})^{T} (\bbw^*-\bbw_{t}) 
		+\ \frac{m}{2} \|\bbw^*-\bbw_{t}\|^2 .
\end{equation}
Observe now that since $\bbw^*$ is the minimizing argument of $F(\bbw)$ we must have  $F(\bbw^*) -\ F(\bbw_{t}) \leq 0$ for all $\bbw$. Using this fact and reordering terms we simplify \eqref{lower_bound_taylor_expansion} to
\begin{equation}\label{eqn_theo_convg_pf_80}
 \frac{m}{2}\  \|\bbw^*-\bbw_{t}\|^2\  \leq\ \nabla F(\bbw_{t})^{T} (\bbw_t-\bbw^*) .
\end{equation}   
Further observe that the Cauchy-Schwarz inequality implies that $\nabla F(\bbw_{t})^{T} (\bbw_{t} - \bbw^*)\leq\|\nabla F(\bbw_{t})\|\|\bbw_{t}-\bbw^{*}\|$. Substitution of this bound in \eqref{eqn_theo_convg_pf_80} and simplification of a $\|\bbw^*-\bbw_{t}\|$ factor yields
\begin{equation}\label{qqqq}
   \frac{m}{2}\| \bbw_{t} -\bbw^{*} \|\  \leq\   \|\nabla F(\bbw_{t})\|.
\end{equation}
Since the limit infimum of $\|\nabla F(\bbw_{t})\|$ is null as stated in \eqref{pppp} the result in \eqref{eqn_convg} follows from considering the bound in \eqref{qqqq} in the limit as the iteration index $t\to\infty$. 

\end{myproof}

%
{Theorem \ref{convg} establishes convergence of the RES algorithm summarized in Algorithm \ref{algo_stochastic_bfgs}. In the proof of the prerequisite Lemma \ref{helpful} the lower bound in the eigenvalues of $\hbB_t$ enforced by the regularization in \eqref{akbar} plays a fundamental role. Roughly speaking, the lower bound in the eigenvalues of $\hbB_t$ results in an upper bound on the eigenvalues of $\hbB_t^{-1}$ which limits the effect of random variations on the stochastic gradient $\hbs(\bbw_{t},\tbtheta_{t})$. If this regularization is not implemented, i.e., if we keep $\delta=0$, we may observe catastrophic amplification of random variations of the stochastic gradient. This effect is indeed observed in the numerical experiments in Section \ref{sec:simulations}. The addition of the identity matrix bias $\Gamma\bbI$ in \eqref{eqn_sbfgs_dual_iteration} is instrumental in the proof of Theorem \ref{convg} proper. This bias limits the effects of randomness in the curvature estimate $\hbB_t$. If random variations in the curvature estimate $\hbB_t$ result in a matrix $\hbB_t^{-1}$ with small eigenvalues the term $\Gamma\bbI$ dominates and \eqref{eqn_sbfgs_dual_iteration} reduces to (regular) SGD. This ensures continued progress towards the optimal argument $\bbw^*$.} 

\subsection{Rate of Convergence} \label{sec:rate}

 {We complement the convergence result in Theorem \ref{convg} with a characterization of the expected convergence rate that we introduce in the following theorem.}
\begin{thm}\label{theo_convergence_rate}
Consider the RES algorithm as defined by \eqref{eqn_sbfgs_dual_iteration}-\eqref{akbar} and let the sequence of step sizes be given by $\epsilon_{t}=\epsilon_{0}T_{0} / (T_{0}+t)$ with the parameter $\eps_0$ sufficiently small and the parameter $T_{0}$ sufficiently large so as to satisfy the inequality
\begin{equation}\label{eqn_thm_cvg_rate_10}
    2\  \epsilon_{0} T_{0} \Gamma >1\ .
\end{equation} 
If assumptions \ref{ass_intantaneous_hessian_bounds},  \ref{ass_bounded_stochastic_gradient_norm} and \ref{ass_delta_less_m} hold true the difference between the expected objective value $\E {F(\bbw_{t})}$ at time $t$ and the optimal objective $F(\bbw^*)$ satisfies
\begin{equation}\label{eqn_thm_cvg_rate_20}
\E {F(\bbw_{t})}- F(\bbw^*)\ \leq\ \frac{C_{0}}{T_{0}+t}\ ,
\end{equation}
where the constant $C_{0}$ satisfies
\begin{equation}\label{eqn_thm_cvg_rate_30}
 C_{0}\ =\ \max\ \left\{ \frac{\epsilon_{0}^{2}\   T_{0}^{2} K}{2 \epsilon_{0} T_{0} \Gamma -1}\ ,\  T_{0}\ \!  (F(\bbw_{0}) -\ F(\bbw^*))  \right\} .
\end{equation}
\end{thm}

%
\begin{myproof} See Appendix. \end{myproof}

Theorem \ref{theo_convergence_rate} shows that under specified assumptions, the expected error  in terms of the objective value after $t$ RES iterations is of order $O(1/t)$. This implies that the rate of convergence for RES is at least linear in expectation. Linear expected convergence rates are typical of stochastic optimization algorithms and, in that sense, no better than conventional SGD. While the convergence rate doesn't change, improvements in convergence time are marked as we illustrate with the numerical experiments of sections \ref{sec:simulations} and \ref{sec:SVM2}.

\section{Numerical analysis}
\label{sec:simulations}

We compare convergence times of RES and SGD in problems with small and large condition numbers. We use a stochastic quadratic objective function as a test case. In particular, consider a positive definite diagonal matrix $\bbA \in \mbS_{n}^{++}$, a vector $\bbb \in \mbR^{n}$, a random vector $\bbtheta\in\reals^n$, and diagonal matrix $\diag(\bbtheta)$ defined by $\bbtheta$. The function $F(\bbw)$ in \eqref{optimization_problem} is defined as
\begin{align}\label{example_problem}
  F(\bbw)  :=&\ \mbE_{\theta}\left[ f(\bbw,{\theta})\right]  \nonumber \\ 
           :=&\ \mbE_{\theta} \left[\frac{1}{2}
               \bbw^{T}\Big(\bbA+\bbA\diag(\bbtheta)\Big)\bbw+\bbb^{T}\bbw\right].
\end{align}
In \eqref{example_problem}, the random vector $\bbtheta$ is chosen uniformly at random from the $n$ dimensional box $\Theta=[-\theta_0,\theta_0]^n$ for some given constant $\theta_0<1$. The linear term $\bbb^{T}\bbw$ is added so that the instantaneous functions $ f(\bbw,{\theta})$ have different minima  which are (almost surely) different from the minimum of the average function $F(\bbw)$. The quadratic term is chosen so that the condition number of $F(\bbw)$ is the condition number of $\bbA$. Indeed, just observe that since $\mbE_{\theta}[\bbtheta]=\bbzero$, the average function in \eqref{example_problem} can be written as $F(\bbw)=(1/2)\bbw^{T}\bbA\bbw+\bbb^{T}\bbw$. The parameter $\theta_0$ controls the variability of the instantaneous functions $f(\bbw,{\theta})$. For small $\theta_0\approx0$ instantaneous functions are close to each other and to the average function. For large $\theta_0\approx1$ instantaneous functions vary over a large range. Further note that we can write the optimum argument as $\bbw^*= \bbA^{-1}\bbb$ for comparison against iterates $\bbw_t$.

For a given $\rho$ we study the convergence metric
\begin{align}\label{eqn_empirical_convergence_time}
   \tau := L \min_t \left\{ t : \frac{\|\bbw_t-\bbw^*\|}{\|\bbw^*\|} \leq \rho\right\},
\end{align}
which represents the time needed to achieve a given relative distance to optimality $\|\bbw_t-\bbw^*\|/\|\bbw^*\|\leq\rho$ as measured in terms of the number $Lt$ of stochastic functions that are processed to achieve such accuracy.


%
\begin{figure}[t]
\centering
\includegraphics[width=\linewidth,height=0.490\linewidth]{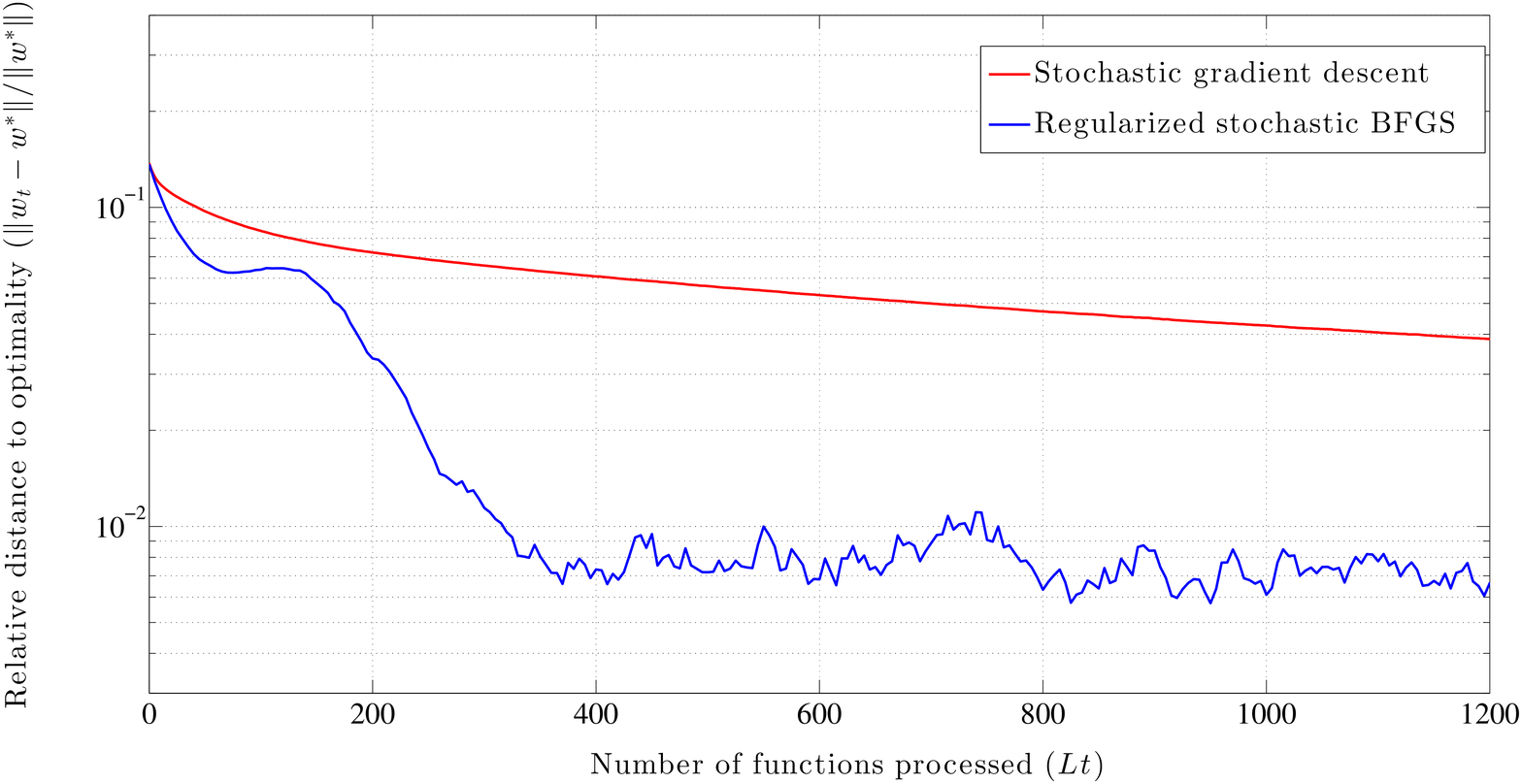}
\caption{ Convergence of SGD and RES for the function in \eqref{example_problem}. Relative distance to optimality $\|\bbw_t-\bbw^*\|/\|\bbw^*\|$ shown with respect to the number $Lt$ of stochastic functions processed. For RES the number of iterations required to achieve a certain accuracy is smaller than the corresponding number for SGD. {See text for parameters values.} }
\label{fig:3}
\end{figure}
%

%
\subsection{Effect of problem's condition number}\label{sec_numerical_analysis_condition_number}

To study the effect of the problem's condition number we generate instances of \eqref{example_problem} by choosing $\bbb$ uniformly at random from the box $[0,1]^n$ and the matrix $\bbA$ as diagonal with elements $a_{ii}$ uniformly drawn from the discrete set $\{1, 10^{-1},\ldots, 10^{-\xi}\}$. This choice of $\bbA$ yields problems with condition number $10^\xi$. 

Representative runs of RES and SGD for $n=50$, $\theta_0=0.5$, and $\xi=2$ are shown in Fig. \ref{fig:3}. For the RES run the stochastic gradients $\hbs(\bbw,\tbtheta)$ in \eqref{stochastic_gradient} are computed as an average of $L=5$ realizations, the regularization parameter in \eqref{jadid} is set to $\delta=10^{-3}$, and the minimum progress parameter in \eqref{eqn_sbfgs_dual_iteration} to $\Gamma=10^{-4}$. For SGD we use $L=1$ in \eqref{stochastic_gradient}. In both cases the step size sequence is of the form $\epsilon_{t}=\epsilon_{0}T_{0}/ (T_{0}+t)$ with  $\epsilon_{0} =10^{-1}$ and $T_{0}=10^3$. Since we are using different value of $L$ for SGD and RES we plot the relative distance to optimality $\|\bbw_t-\bbw^*\|/\|\bbw^*\|$ against the number $Lt$ of functions processed up until iteration $t$.

As expected for a problem with a large condition number RES is much faster than SGD. After $t=1,200$ the distance to optimality for the SGD iterate is {$\|\bbw_t-\bbw^*\|/\|\bbw^*\|=3.8\times10^{-2}$}. Comparable accuracy $\|\bbw_t-\bbw^*\|/\|\bbw^*\|=3.8\times10^{-2}$ for RES is achieved after $t=38$ iterations. Since we are using $L=5$ for RES this corresponds to $Lt=190$ random function evaluations. Conversely, upon processing  $Lt=1,200$ random functions -- which corresponds to $t=240$ iterations -- RES achieves accuracy $\|\bbw_t-\bbw^*\|/\|\bbw^*\|=6.6\times10^{-3}$. This relative performance difference can be made arbitrarily large by modifying the condition number of $\bbA$. 

%
\begin{figure}[t]
\centering
\includegraphics[width=\linewidth,height=0.490\linewidth]{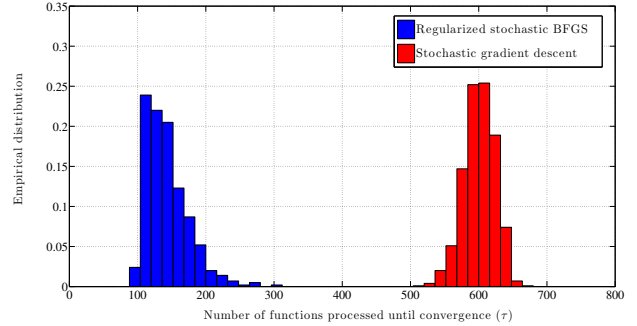}
\caption{ Convergence of SGD and RES for well conditioned problems. Empirical distributions of the number $\tau=Lt$ of stochastic functions that are processed to achieve relative precision $\|\bbw_t-\bbw^*\|/\|\bbw^*\|\leq10^{-2}$ are shown. Histogram is across {$J=1,000$} realizations of functions as in \eqref{example_problem} with condition number $10^\xi=1$. Convergence for RES is better than SGD but the number of iterations required for convergence is of the same order of magnitude. See text for parameters values.
}
\label{fig:1}
\end{figure}

%
\begin{figure}[t]
\centering
\includegraphics[width=\linewidth,height=0.490\linewidth]{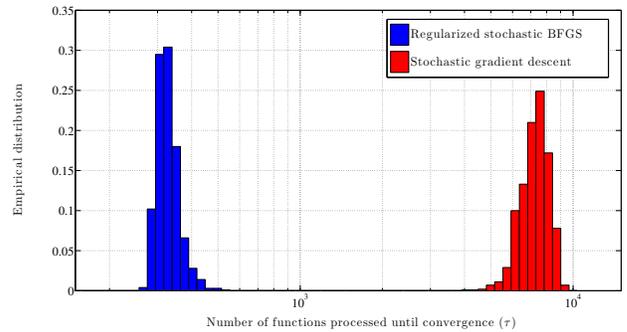}
 \caption{ Convergence of SGD and RES for ill conditioned problems. Empirical distributions of the number $\tau=Lt$ of stochastic functions that are processed to achieve relative precision $\|\bbw_t-\bbw^*\|/\|\bbw^*\|\leq10^{-2}$ are shown. Histogram is across {$J=1,000$} realizations of functions as in \eqref{example_problem} with condition number $10^\xi=10^2$. RES reduces the convergence time of SGD by an order of magnitude. See text for parameters values.
 }
\label{fig:2}
\end{figure}

A more comprehensive analysis of the relative advantages of RES appears in figs. \ref{fig:1} and \ref{fig:2}. We keep the same parameters used to generate Fig. \ref{fig:3} except that we use $\xi=0$ for Fig. \ref{fig:1} and $\xi=2$ for Fig. \ref{fig:2}. This yields a family of well-condition functions with condition number $10^\xi=1$ and a family of ill-conditioned functions with condition number  $10^\xi=10^2$. In both figures we consider $\rho=10^{-2}$ and study the convergence times $\tau$ and $\tau'$ of RES and SGD, respectively [cf. \eqref{eqn_empirical_convergence_time}]. Resulting empirical distributions of $\tau$ and $\tau'$ across {$J=1,000$ instances of the functions $F(\bbw)$ in \eqref{example_problem}} are reported in figs. \ref{fig:1} and \ref{fig:2} for the well conditioned and ill conditioned families, respectively. For the well conditioned family RES reduces the number of functions processed from an average of {$\bar\tau'=601$} in the case of SGD to an average of {$\bar\tau=144$}. This nondramatic improvement becomes more significant for the ill conditioned family where the reduction is from an average of {$\bar\tau'=7.2 \times10^3$} for SGD to an average of {$\bar\tau=3.2 \times10^2$} for RES. The spread in convergence times is also smaller for RES.

%
\begin{figure}[t]
\centering
\includegraphics[width=\linewidth,height=0.490\linewidth]{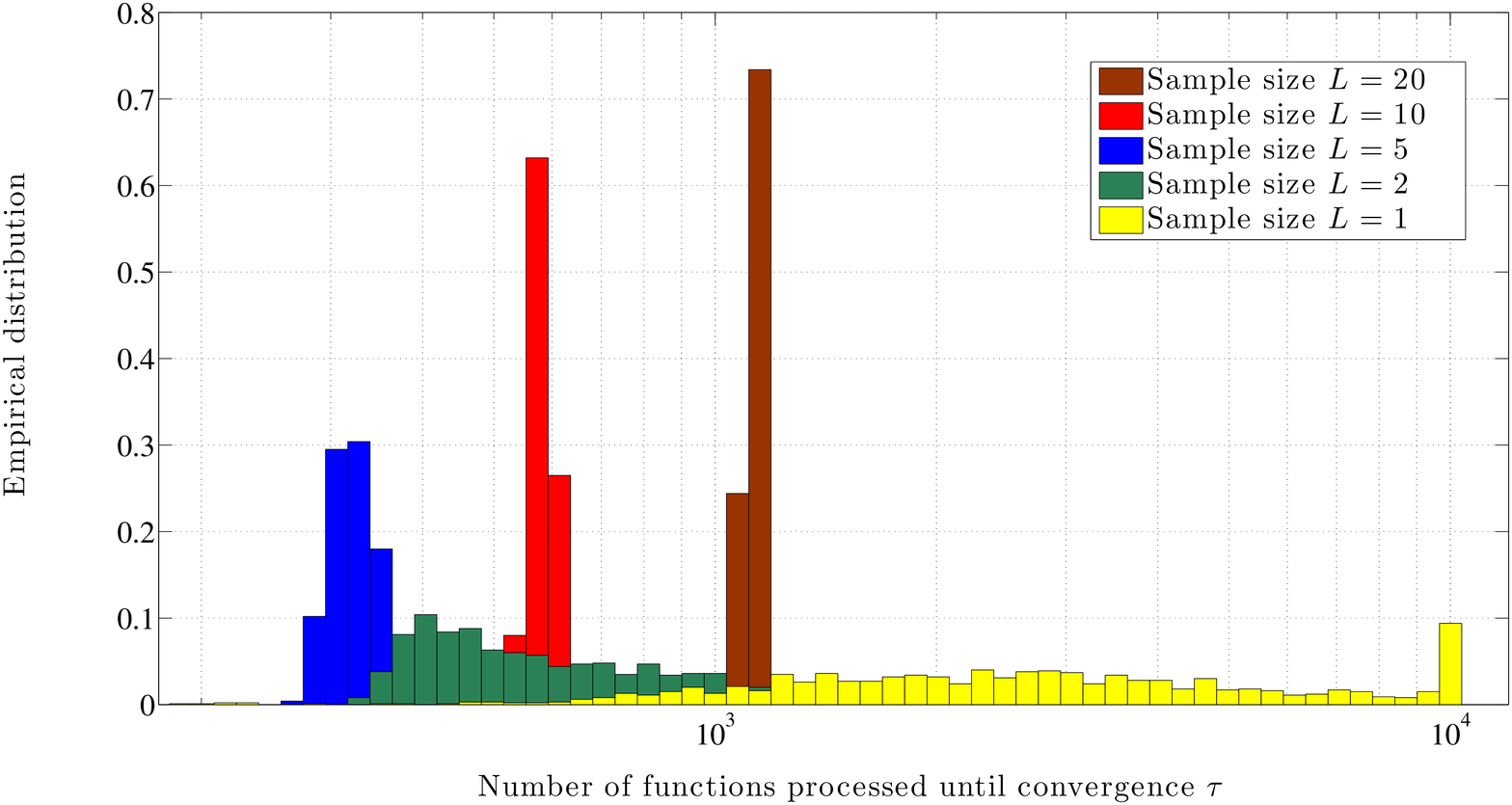}
\caption{ Convergence of RES for different sample sizes in the computation of stochastic gradients. Empirical distributions of the number $\tau=Lt$ of stochastic functions that are processed to achieve relative precision $\|\bbw_t-\bbw^*\|/\|\bbw^*\|\leq10^{-2}$ are shown when we use $L=1$, $L=2$, $L=5$, $L=10$, and $L=20$ in the evaluation of the stochastic gradients $\hbs(\bbw,\tbtheta)$ in \eqref{stochastic_gradient}. The average convergence time decreases as we go from small to moderate values of $L$ and starts increasing as we go from moderate to large values of $L$. The variance of convergence times decreases monotonically with increasing $L$. See text for parameters values.}
\label{fig:SSE}
\end{figure}

%
\subsection{Choice of stochastic gradient average}

{The stochastic gradients $\hbs(\bbw,\tbtheta)$ in \eqref{stochastic_gradient} are computed as an average of $L$ sample gradients $\nabla f(\bbw,{\bbtheta_{l}})$. To study the effect of the choice of $L$ on RES we consider problems as in \eqref{example_problem} with matrices $\bbA$ and vectors $\bbb$ generated as in Section \ref{sec_numerical_analysis_condition_number}.} {We consider problems with $n=50$, $\theta_0=0.5$, and $\xi=2$; set the RES parameters to $\delta=10^{-3}$ and $\Gamma=10^{-4}$; and the step size sequence to $\epsilon_{t}=\epsilon_{0}T_{0} / (T_{0}+t)$ with  $\epsilon_{0} =10^{-1}$ and $T_{0}=10^3$. We then consider different choices of $L$ and for each specific value generate {$J=1,000$} problem instances. For each run we record the total number $\tau_L$ of sample functions that need to be processed to achieve relative distance to optimality $\|\bbw_t-\bbw^*\|/\|\bbw^*\| \leq10^{-2}$ [cf. \eqref{eqn_empirical_convergence_time}].  If $\tau>10^4$ we report $\tau=10^4$ and interpret this outcome as a convergence failure. The resulting estimates of the probability distributions of the times $\tau_L$ are reported in Fig. \ref{fig:SSE} for $L=1$, $L=2$, $L=5$, $L=10$, and $L=20$.} 

{The trends in convergence times $\tau$ apparent in Fig. \ref{fig:SSE} are: (i) As we increase $L$ the variance of convergence times decreases. (ii) The average convergence time decreases as we go from small to moderate values of $L$ and starts increasing as we go from moderate to large values of $L$.}
 {Indeed, the empirical standard deviations of convergence times decrease monotonically from $\sigma_{\tau_1} = 2.8 \times 10^{3}$ to $\sigma_{\tau_2} = 2.6 \times 10^{2}$, $\sigma_{\tau_5} = 31.7$, $\sigma_{\tau_{10}} = 28.8$, and $\sigma_{\tau_{20}} = 22.7$, when $L$ increases from $L=1$ to $L=2$, $L=5$, $L=10$, and $L=20$. The empirical mean decreases from $\bar\tau_{1} =3.5 \times 10^{3} $ to $\bar\tau_{2} = 6.3 \times 10^{2}$ as we move from $L=1$ to $L=2$, stays at about the same value $\bar\tau_{5} = 3.3 \times 10^{2}$ for $L=5$ and then increases to $\bar\tau_{10} = 5.8 \times 10^{2} $ and $\bar\tau_{20} =1.2 \times 10^{3} $ for $L=10$ and $L=20$.} {This behavior is expected since increasing $L$ results in curvature estimates $\hbB_t$ closer to the Hessian $\bbH(\bbw_t)$ thereby yielding better convergence times. As we keep increasing $L$, there is no payoff in terms of better curvature estimates and we just pay a penalty in terms of more function evaluations for an equally good $\hbB_t$ matrix. This can be corroborated by observing that the convergence times $\tau_{5}$ are about half those of $\tau_{10}$ which in turn are about half those of $\tau_{20}$. This means that the {\it actual} convergence times $\tau/L$ have similar distributions for $L=5$, $L=10$, and $L=20$.} {The empirical distributions in Fig. \ref{fig:SSE} show that moderate values of $L$ suffice to provide workable curvature approximations. This justifies the use $L=5$ in sections \ref{sec_numerical_analysis_condition_number} and \ref{sec_numerical_analysis_problem_dimension}}
 

%

\begin{figure*}[t]
        \centering
        \begin{subfigure}[b]{0.5\textwidth}
                \includegraphics[width=\textwidth]{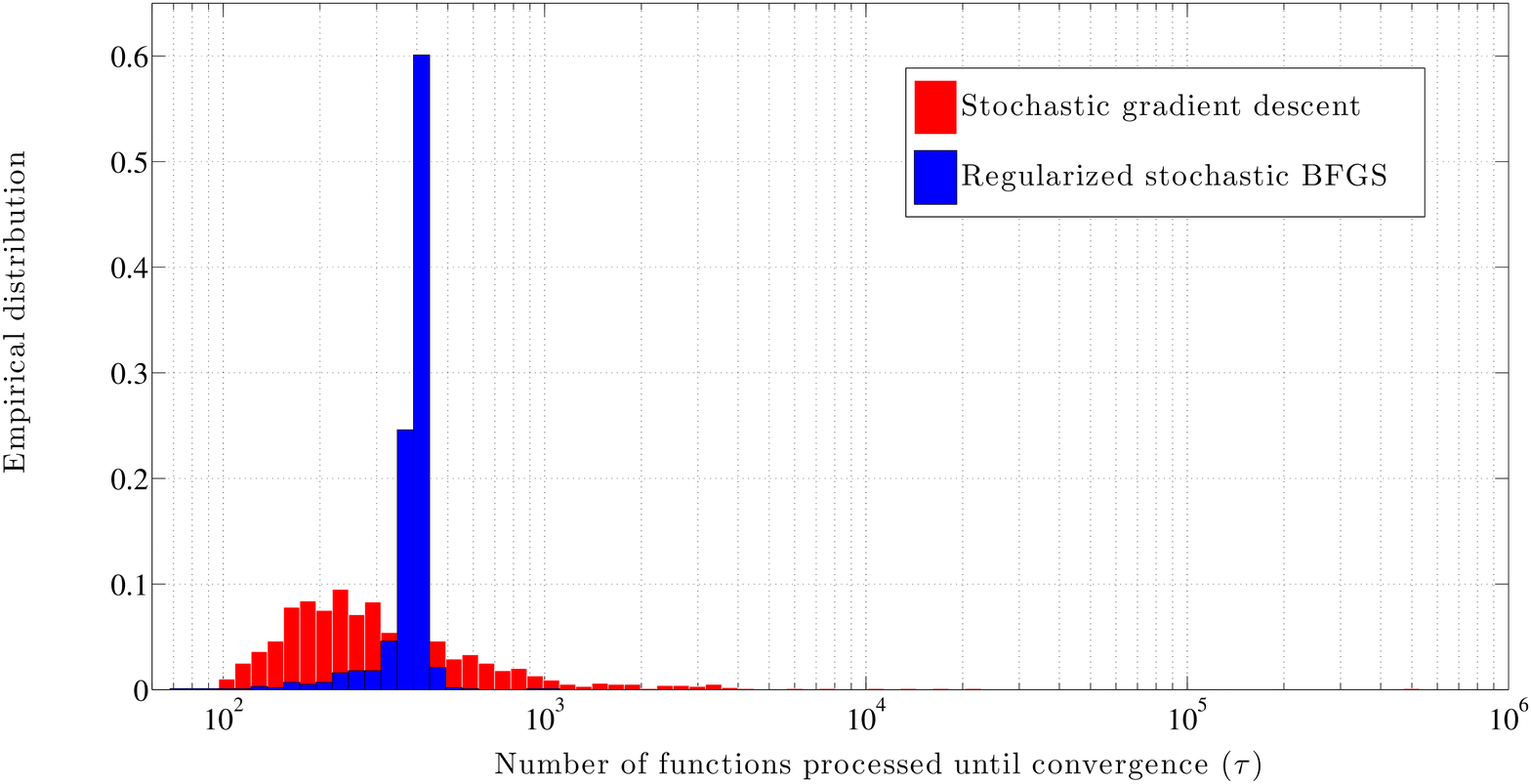}
                \caption{$n = 5$}
                \label{fig:n=5}
        \end{subfigure}%
        ~ 
        \begin{subfigure}[b]{0.5\textwidth}
                \includegraphics[width=\textwidth]{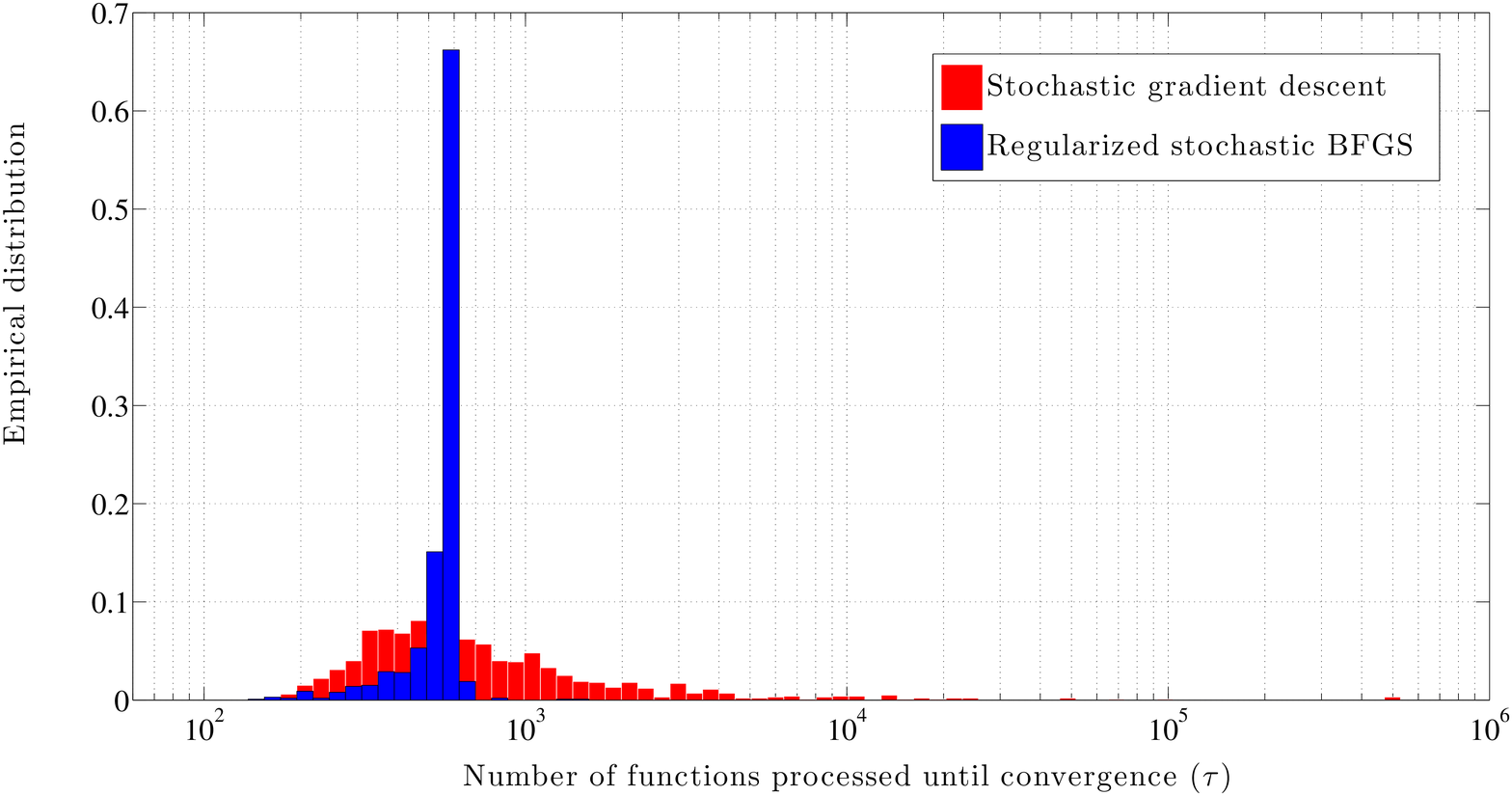}
                \caption{$n=10$}
                \label{fig:n=10}
        \end{subfigure}
        ~ 
        \begin{subfigure}[b]{0.5\textwidth}
                \includegraphics[width=\textwidth]{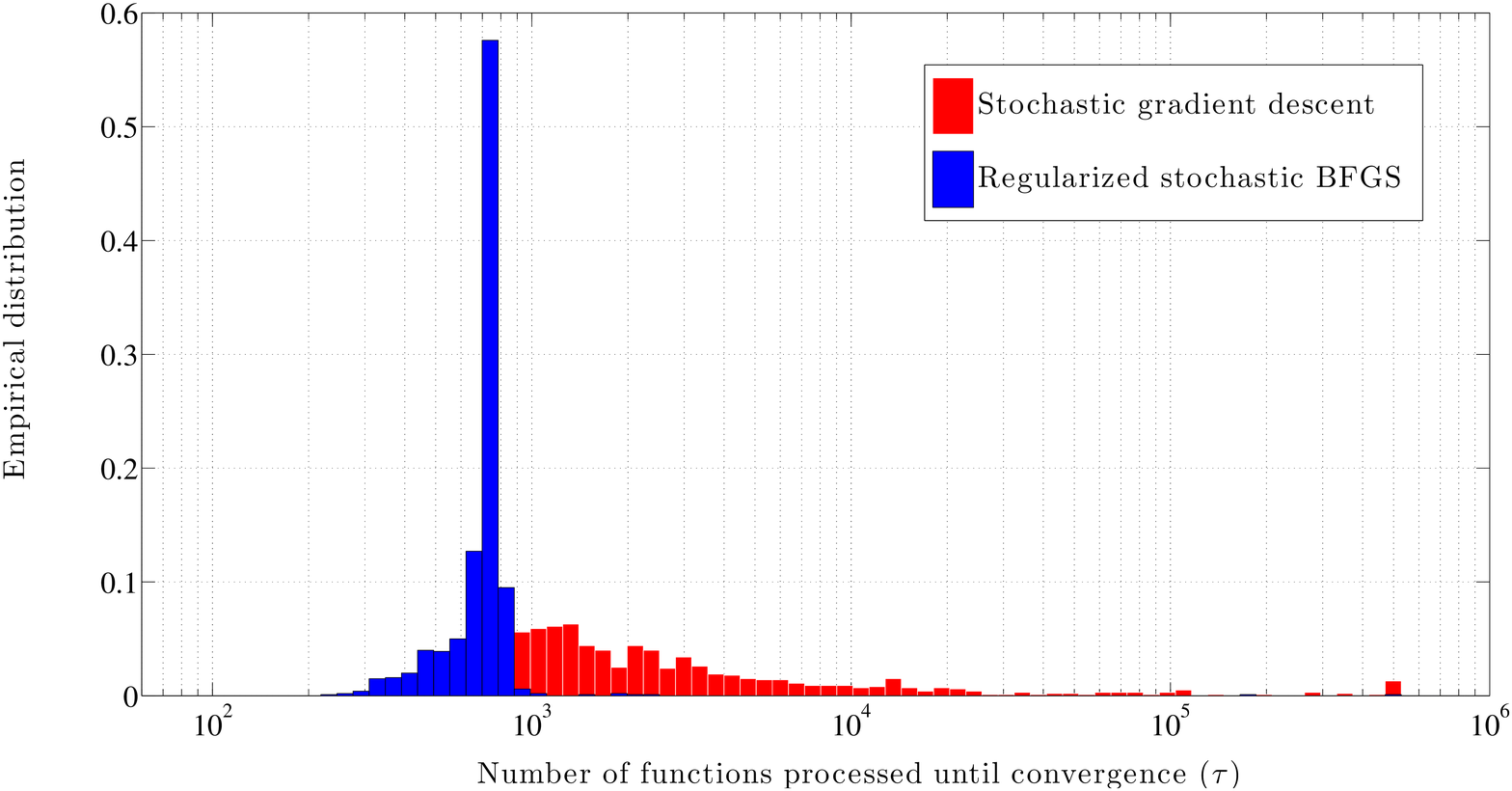}
                \caption{$n=20$}
                \label{fig:n=20}
        \end{subfigure}%
           ~ 
        \begin{subfigure}[b]{0.5\textwidth}
                \includegraphics[width=\textwidth]{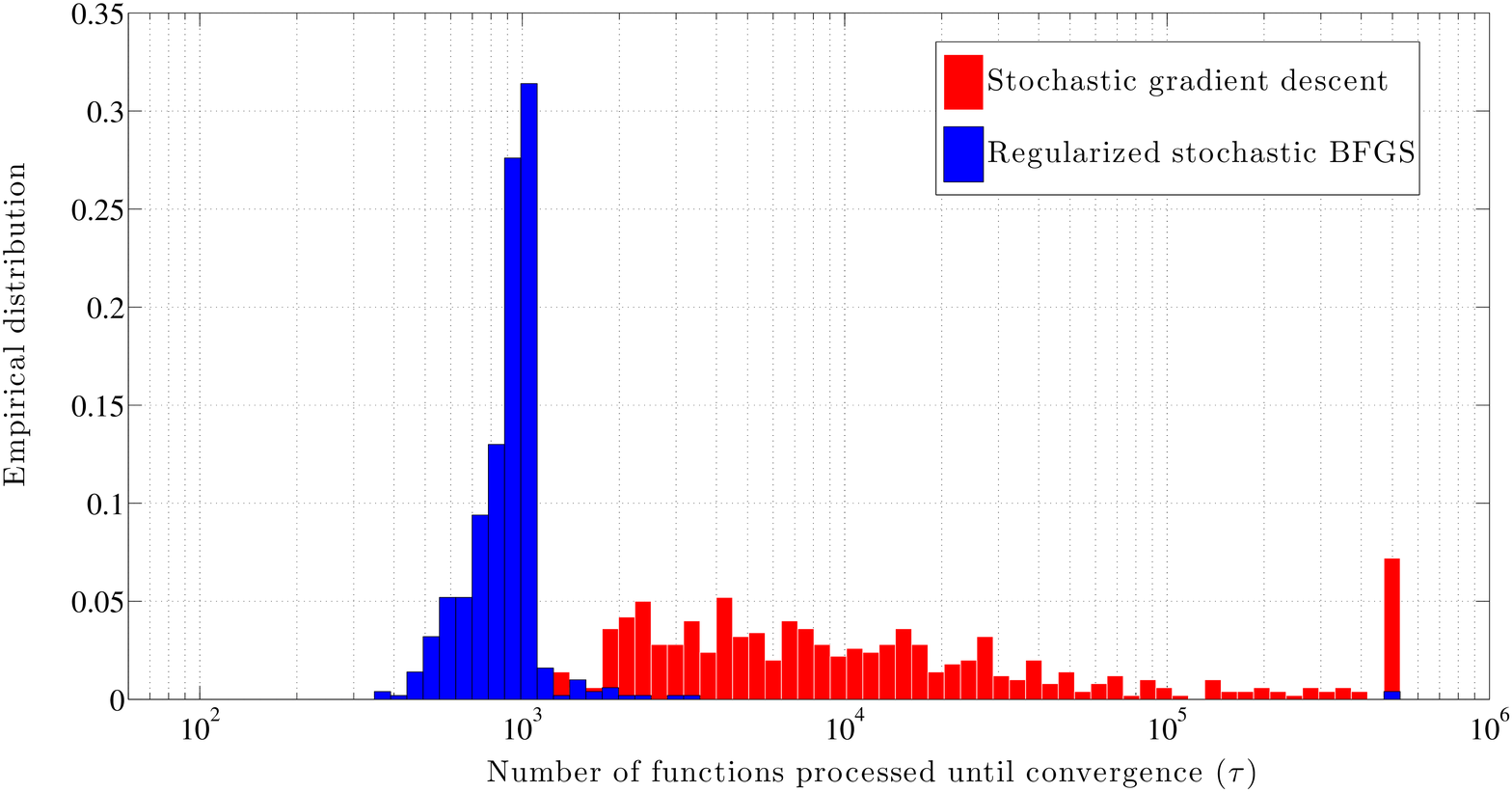}
                \caption{$n=50$}
                \label{fig:n=50}
        \end{subfigure}
        \caption{ Histogram of the number of data points that SGD and RES needs to converge. Convergence time for RES increases smoothly by increasing the dimension of problem, while convergence time of SGD increases faster.  }\label{fig:dimension}
\end{figure*}

%
\subsection{Effect of problem's dimension}\label{sec_numerical_analysis_problem_dimension}

To evaluate performance for problems of different dimensions we consider functions of the form in \eqref{example_problem} with $\bbb$ uniformly chosen from the box $[0,1]^n$ and diagonal matrix $\bbA$ as in Section \ref{sec_numerical_analysis_condition_number}. However, we select the elements $a_{ii}$ as uniformly drawn from the interval $[0,1]$. This results in problems with more moderate condition numbers and allows for a comparative study of performance degradations of RES and SGD as the problem dimension $n$ grows. 

{The variability parameter for the random vector $\bbtheta$ is set to $\theta_0=0.5$. The RES parameters are $L=5$, $\delta=10^{-3}$, and $\Gamma=10^{-4}$. For SGD we use $L=1$. In both methods the step size sequence is $\epsilon_{t}=\epsilon_{0}T_{0} / (T_{0}+t)$ with  $\epsilon_{0} =10^{-1}$ and $T_{0}=10^3$. For a problem of dimension $n$ we study convergence times $\tau_n$ and $\tau'_n$ of RES and SGD as defined in \eqref{eqn_empirical_convergence_time} with $\rho=1$. For each value of $n$ considered we determine empirical distributions of $\tau_n$ and $\tau'_n$ across $J=1,000$ problem instances. If $\tau>5 \times 10^5$ we report $\tau= 5 \times 10^5$ and interpret this outcome as a convergence failure. The resulting histograms are shown in Fig. \ref{fig:dimension} for $n=5$, $n=10$, $n=20$, and $n=50$.} 

 {For problems of small dimension having $n=5$ the average performances of RES and SGD are comparable, with SGD performing slightly better. E.g., the medians of these times are $\text{median}(\tau_5)=400$ and $\text{median}(\tau'_5)=265$, respectively. A more significant difference is that times $\tau_5$ of RES are more concentrated than times $\tau'_5$ of SGD. The latter exhibits large convergence times $\tau'_5>10^3$ with probability {$0.06$} and fails to converge altogether in a few rare instances -- we have $\tau'_5=5\times10^5$ in {1 out of 1,000 realizations}. In the case of RES all realizations of $\tau_5$ are in the interval {$70\leq\tau_5\leq1095$}}. 

As we increase $n$ we see that RES retains the smaller spread advantage while eventually exhibiting better average performance as well. Medians for $n=10$ are still comparable at {$\text{median}(\tau_{10})=575$ and $\text{median}(\tau'_{10})=582$}, as well as for $n=20$ at {$\text{median}(\tau_{20})=745$ and $\text{median}(\tau'_{20})=1427$}. For $n=50$ the RES median is decidedly better since {$\text{median}(\tau_{50})=950$ and $\text{median}(\tau'_{50})=7942$}. 

For large dimensional problems having $n=50$ SGD becomes unworkable. It fails to achieve convergence in $5 \times 10^5$ iterations with probability $0.07$ and exceeds $10^4$ iterations {with probability $0.45$}. For RES we fail to achieve convergence in $5\times 10^5$ iterations with probability $3\times10^{-3}$ and achieve convergence in less than $10^4$ iterations in all other cases. Further observe that RES degrades smoothly as $n$ increases. The median number of gradient evaluations needed to achieve convergence increases by a factor of {$\text{median}(\tau'_{50})/\text{median}(\tau'_{5}) = 29.9$} as we increase $n$ by a factor of $10$. The spread in convergence times remains stable as $n$ grows.

\section{Support vector machines} 
\label{sec:SVMproblem}

A particular case of \eqref{optimization_problem} is the implementation of a support vector machine (SVM). Given a training set with points whose class is known the goal of a SVM is to find a hyperplane that best separates the training set. To be specific let $\ccalS = \{ (\bbx_{i},y_{i}) \}_{i=1}^{N}$ be a training set containing $N$ pairs of the form $(\bbx_{i},y_i)$, where $\bbx_{i}\in\reals^n$ is a feature vector and $y_{i}\in \{-1,1 \}$ is the corresponding vector's class. The goal is to find a hyperplane supported by a vector $\bbw\in\reals^n$ which separates the training set so that $\bbw^T\bbx_i>0$ for all points with $y_i=1$ and $\bbw^T\bbx_i<0$ for all points with $y_i=-1$. This vector may not exist if the data is not perfectly separable, or, if the data is separable there may be more than one separating vector. We can deal with both situations with the introduction of a loss function $l((\bbx,y);\bbw)$ defining some measure of distance between the point $\bbx_i$ and the hyperplane supported by $\bbw$. We then select the hyperplane supporting vector as
\begin{equation}\label{SVM}
   \bbw^* := \argmin_{\bbw}\   \frac{\lambda}{2}\|\bbw\|^2 
                      + \frac{1}{N} \sum_{i=1}^{N} l((\bbx_{i},y_{i});\bbw),
\end{equation}
where we also added the regularization term ${\lambda}\|\bbw\|^2 /{2} $ for some constant $\lambda>0$. The vector $\bbw^*$ in \eqref{SVM} balances the minimization of the sum of distances to the separating hyperplane, as measured by the loss function $l((\bbx,y);\bbw)$, with the minimization of the $L_{2}$ norm $\|\bbw\|_2$ to enforce desirable properties in $\bbw^*$. Common selections for the loss function are the hinge loss $l((\bbx,y);\bbw)=\max(0,1-y(\bbw^{T}\bbx))$, the squared hinge loss $l((\bbx,y);\bbw)=\max(0,1-y(\bbw^{T}\bbx))^{2}$ and the log loss $l((\bbx,y);\bbw)=\log(1+\exp(-y(\bbw^{T}\bbx)))$. See, e.g., \cite{Vapnik, Bottou}. 

In order to model \eqref{SVM} as a stochastic optimization problem in the form of problem \eqref{optimization_problem}, we define $\bbtheta_{i}=(\bbx_{i},y_{i})$ as a given training point and $m_{\bbtheta}(\bbtheta)$ as a uniform probability distribution on the training set $\ccalS = \{ (\bbx_{i},y_{i}) \}_{i=1}^{N}= \{ \bbtheta_{i} \}_{i=1}^{N}$. Upon defining the sample functions 
\begin{equation}\label{eqn_svn_reformulation_random_functions}
f(\bbw,\bbtheta)  =\ f(\bbw,(\bbx,y))\ 
                 :=\ \frac{\lambda}{2}\|\bbw\|^2 + l((\bbx,y);\bbw),
\end{equation}
it follows that we can rewrite the objective function in \eqref{SVM} as
\begin{equation}\label{eqn_svn_reformulation}
  \frac{\lambda}{2}\|\bbw\|^2 
        + \frac{1}{N} \sum_{i=1}^{N} l((\bbx_{i},y_{i});\bbw)  
    \ =\ \mbE_{\bbtheta} [ f(\bbw,\bbtheta)]
\end{equation}
since each of the functions $f(\bbw,\bbtheta)$ is drawn with probability $1/N$ according to the definition of $m_{\bbtheta}(\bbtheta)$. Substituting \eqref{eqn_svn_reformulation} into \eqref{SVM} yields a problem with the general form of \eqref{optimization_problem} with random functions $f(\bbw,\bbtheta)$ explicitly given by \eqref{eqn_svn_reformulation_random_functions}.

We can then use Algorithm \eqref{algo_stochastic_bfgs} to attempt solution of \eqref{SVM}. For that purpose we particularize Step 2 to the drawing of $L$ feature vectors $\tbx_{t}=  [\bbx_{t1}; \dots;\bbx_{tL} ]$ and their corresponding class values $\tby_{t}=  [y_{t1}; \dots;y_{tL}]$ to construct the vector of pairs $\tbtheta_{t}=  [(\bbx_{t1},y_{t1}); \dots;(\bbx_{tL},y_{tL}) ]$ . These training points are selected uniformly at random from the training set $\ccalS$. We also need to particularize steps 3 and 5 to evaluate the stochastic gradient of the specific instantaneous function in \eqref{eqn_svn_reformulation_random_functions}. E.g., Step 3 takes the form
\begin{align}\label{eqn_svn_algorithm_step_3}
  \hbs(\bbw_{t},\tbtheta_t)
     &\ =\ \hbs(\bbw_{t},(\tbx_{t},\tby_{t}))  \nonumber\\
     &\ =\  \lambda\bbw_{t} 
            + \frac{1}{L} \sum_{i=1}^{L} \nabla_{\bbw}\ 
                  \l((\bbx_{ti},y_{ti});\bbw_{t}).
\end{align}
The specific form of Step 5 is obtained by replacing $\bbw_{t+1}$ for $\bbw_{t}$ in \eqref{eqn_svn_algorithm_step_3}. We analyze the behavior of Algorithm \eqref{optimization_problem} in the implementation of a SVM in the following section.

\subsection{Numerical Analysis}\label{sec:SVM2}

We test Algorithm 1 when using the squared hinge loss $l((\bbx,y);\bbw)=\max(0,1-y(\bbx^{T}\bbw))^{2}$ in \eqref{SVM}. The training set $\ccalS = \{ (\bbx_{i},y_{i}) \}_{i=1}^{N}$ contains $N=10^4$ feature vectors half of which belong to the class $y_i=-1$ with the other half belonging to the class $y_i=1$. For the class $y_i=-1$ each of the $n$ components of each of the feature vectors $\bbx_i\in\reals^n$ is chosen uniformly at random from the interval $[-0.8,0.2]$. Likewise, each of the $n$ components of each of the feature vectors $\bbx_i\in\reals^n$ is chosen uniformly at random from the interval $[-0.2,0.8]$ for the class $y_i=1$. The overlap in the range of the feature vectors is such that the classification accuracy expected from a clairvoyant classifier that knows the statistic model of the data set is less than $100\%$. Exact values can be computed from the Irwin-Hall distribution \cite{Johnson}. For $n=4$ this amounts to $98\%$.

In all of our numerical experiments the parameter $\lambda$ in \eqref{SVM} is set to $\lambda=10^{-3}$. Recall that since the Hessian eigenvalues of $f(\bbw,\bbtheta) :=\lambda\|\bbw\|^2/2 + l((\bbx_{i},y_{i});\bbw)$ are, at least, equal to $\lambda$ this implies that the eigenvalue lower bound $\tdm$ is such that $\tdm\geq\lambda=10^{-3}$. We therefore set the RES regularization parameter to $\delta=\lambda=10^{-3}$. Further set the minimum progress parameter in \eqref{stochastic_gradient} to $\Gamma=10^{-4}$ and the sample size for computation of stochastic gradients to $L=5$. The stepsizes are of the form $\epsilon_t=\epsilon_{0}T_{0} / (T_{0}+t)$ with $\epsilon_{0} = 3\times10^{-2}$ and $T_{0}=10^3$. We compare the behavior of SGD and RES for a small dimensional problem with $n=4$ and a large dimensional problem with $n=40$. For SGD the sample size in \eqref{stochastic_gradient} is $L=1$ and we use the same stepsize sequence used for RES.

An illustration of the relative performances of SGD and RES for $n\!=4$ is presented in Fig. \ref{fig:n=4}. The value of the objective function $F(\bbw_t)$ is represented with respect to the number of feature vectors processed, which is given by the product $Lt$ between the iteration index and the sample size used to compute stochastic gradients. This is done because the sample sizes in RES ($L=5$) and SGD ($L=1$) are different. The curvature correction of RES results in significant reductions in convergence time. E.g., RES achieves an objective value of $F(\bbw_t)=6.5\times10^{-2}$ upon processing of $Lt=315$ feature vectors. To achieve the same objective value $F(\bbw_t)=6.5\times10^{-2}$ SGD processes {$1.74\times10^3$} feature vectors. Conversely, after processing $Lt=2.5\times10^3$ feature vectors the objective values achieved by RES and SGD are $F(\bbw_t) =4.14\times10^{-2}$ and $F(\bbw_t)=6.31\times10^{-2}$, respectively. 

The performance difference between the two methods is larger for feature vectors of larger dimension $n$. The plot of the value of the objective function $F(\bbw_t)$ with respect to the number of feature vectors processed $Lt$ is shown in Fig. \ref{fig:n=40} for $n=40$. The convergence time of RES increases but is still acceptable. For SGD the algorithm becomes unworkable. After processing $3.5\times 10^{3}$ RES reduces the objective value to  $F(\bbw_t)=5.55\times10^{-4}$ while SGD has barely made progress at $F(\bbw_t)=1.80\times10^{-2}$.

Differences in convergence times translate into differences in classification accuracy when we process all $N$ vectors in the training set. This is shown for dimension $n=4$ and training set size $N=2.5\times10^3$ in Fig. \ref{accuracy}. To build Fig. \ref{accuracy} we process $N=2.5\times10^3$ feature vectors with RES and SGD with the same parameters used in Fig. \ref{fig:n=4}. We then use these vectors to classify $10^4$ observations in the test set and record the percentage of samples that are correctly classified. The process is repeated $10^3$ times to estimate the probability distribution of the correct classification percentage represented by the histograms shown. The dominance of RES with respect to SGD is almost uniform. The vector $\bbw_t$ computed by SGD classifies correctly at most $65\%$ of the of the feature vectors in the test set. The vector $\bbw_t$ computed by RES exceeds this accuracy with probability $0.98$. Perhaps more relevant, the classifier computed by RES achieves a mean classification accuracy of $82.2\%$ which is not far from the clairvoyant classification accuracy of $98\%$. Although performance is markedly better in general, RES fails to compute a working classifier with probability $0.02$. We omit comparison of classification accuracy for $n=40$ due to space considerations. As suggested by Fig. \ref{fig:n=40} the differences are more significant than for the case $n=4$.

We also investigate the difference between regularized and non-regularized versions of stochastic BFGS for feature vectors of dimension $n=10$. Observe that non-regularized stochastic BFGS corresponds to making $\delta=0$ and $\Gamma=0$ in Algorithm 1. To illustrate the advantage of the regularization induced by the proximity requirement in \eqref{jadid}, as opposed to the non regularized proximity requirement in \eqref{jaygozin}, we keep a constant stepsize $\epsilon_{t}=10^{-1}$. The corresponding evolutions of the objective function values $F(\bbw_t)$ with respect to the number of feature vectors processed $Lt$ are shown in Fig. \ref{jumps} along with the values associated with stochastic gradient descent. As we reach convergence the likelihood of having small eigenvalues appearing in $\hbB_t$ becomes significant. In regularized stochastic BFGS (RES) this results in recurrent jumps away from the optimal classifier $\bbw^*$. However, the regularization term limits the size of the jumps and further permits the algorithm to consistently recover a reasonable curvature estimate. In Fig. \ref{jumps} we process $10^4$ feature vectors and observe many occurrences of small eigenvalues. However, the algorithm always recovers and heads back to a good approximation of $\bbw^*$. In the absence of regularization small eigenvalues in $\hbB_t$ result in larger jumps away from $\bbw^*$. This not only sets back the algorithm by a much larger amount than in the regularized case but also results in a catastrophic deterioration of the curvature approximation matrix $\hbB_t$. In Fig. \ref{jumps} we observe recovery after the first two occurrences of small eigenvalues but eventually there is a catastrophic deviation after which non-regularized stochastic BFSG behaves not better than SGD.

\begin{figure}[t]
\centering
\includegraphics[width=\linewidth,height=0.48\linewidth]{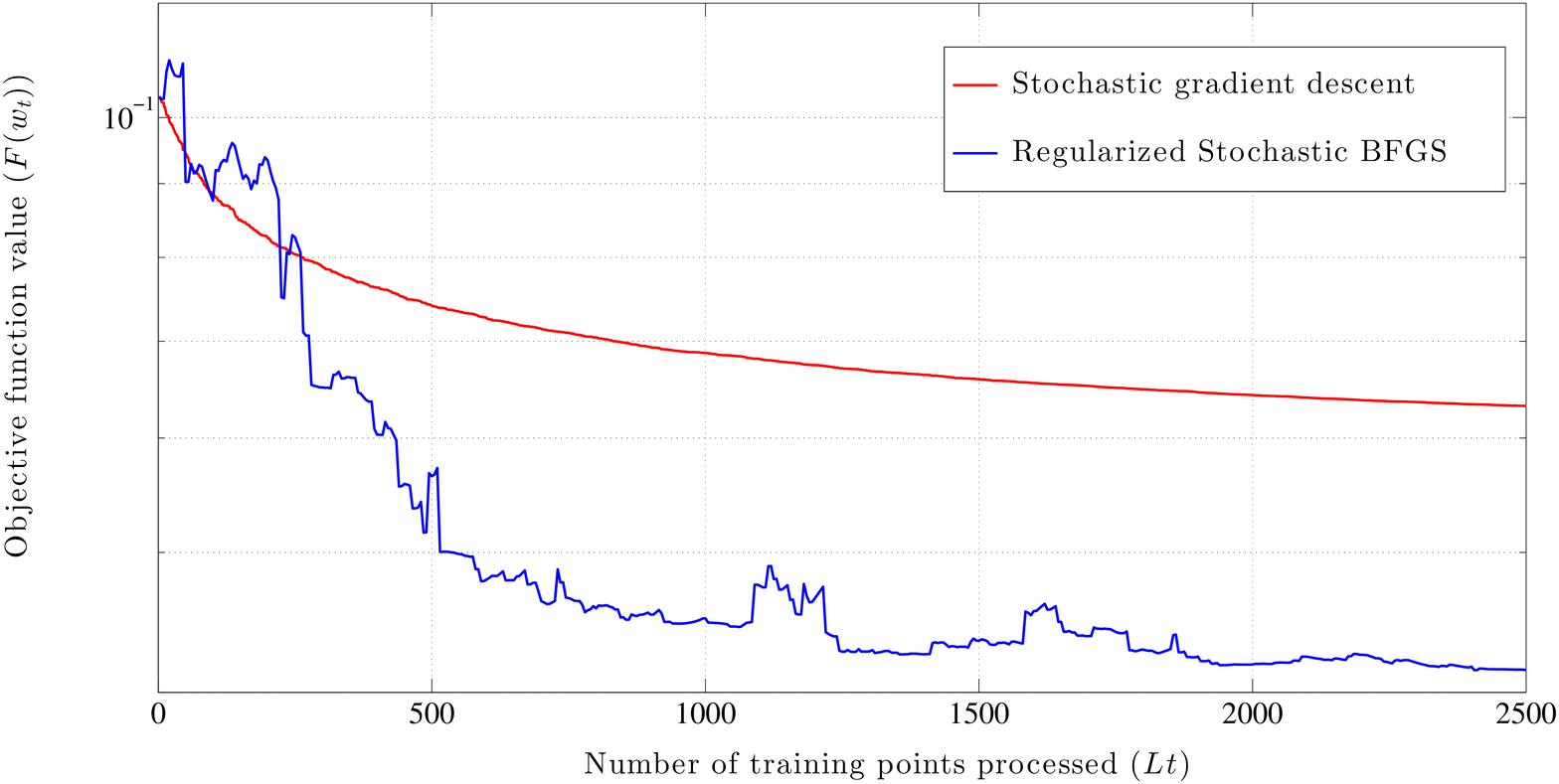}
\caption{Convergence of SGD and RES for feature vectors of dimension $n=4$. Convergence of RES is faster than convergence of SGD (RES sample size $L=5$; SGD sample size $L=1$; stepsizes $\epsilon_t=\epsilon_{0}T_{0} / (T_{0}+t)$ with $\epsilon_{0} = 3\times10^{-2}$ and $T_{0}=10^3$; RES parameters  $\delta=10^{-3}$ and $\Gamma=10^{-4}$).}
\label{fig:n=4}
\end{figure}

\begin{figure}[t]
\centering
\includegraphics[width=\linewidth,height=0.48\linewidth]{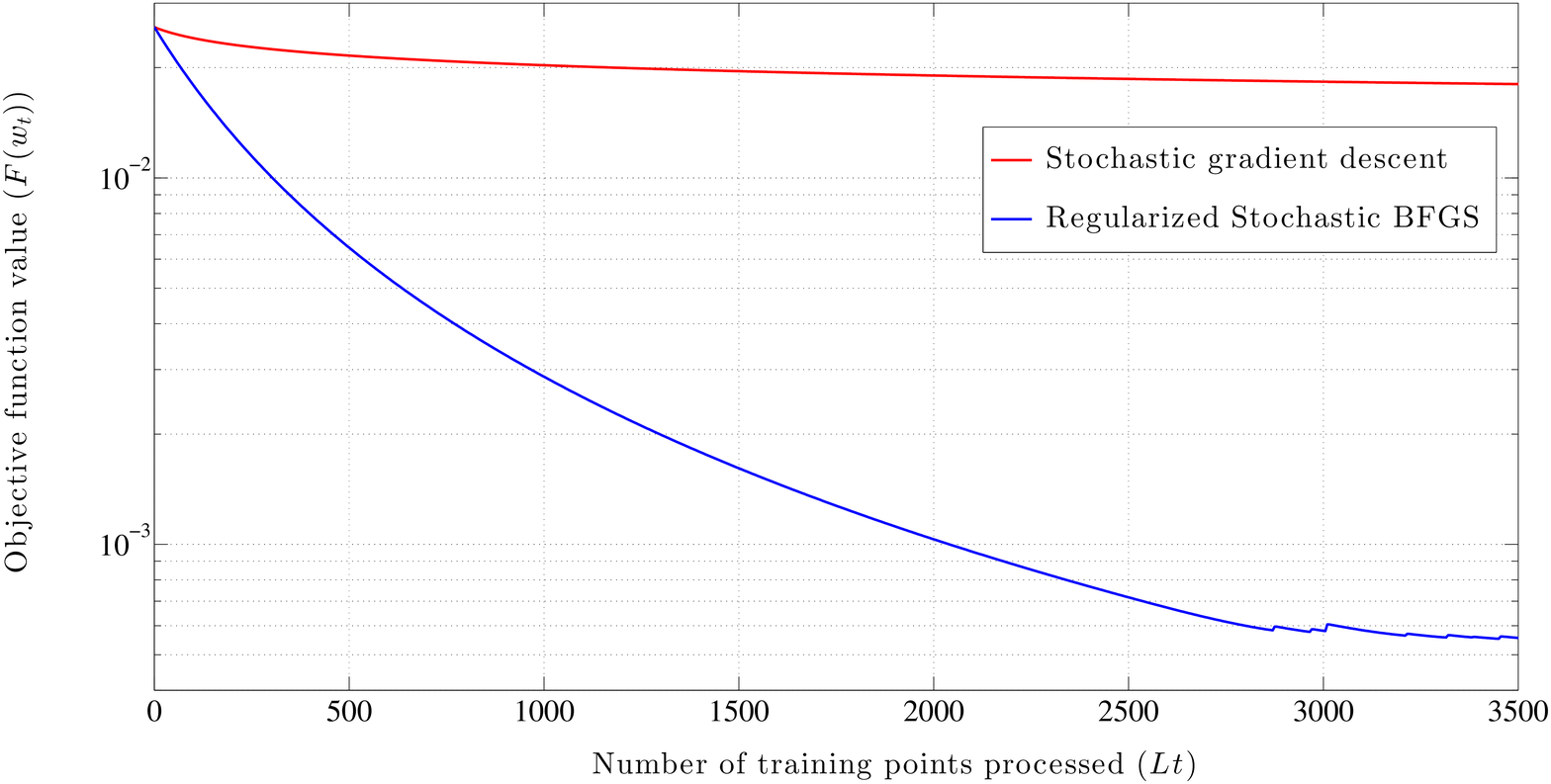}
\caption{Convergence of SGD and RES for feature vectors of dimension $n=40$. RES is still practicable whereas SGD becomes too slow for practical use (parameters are as in Fig. \ref{fig:n=4}).}
\label{fig:n=40}
\end{figure}

\begin{figure}[t]
\centering
\includegraphics[width=\linewidth,height=0.48\linewidth]{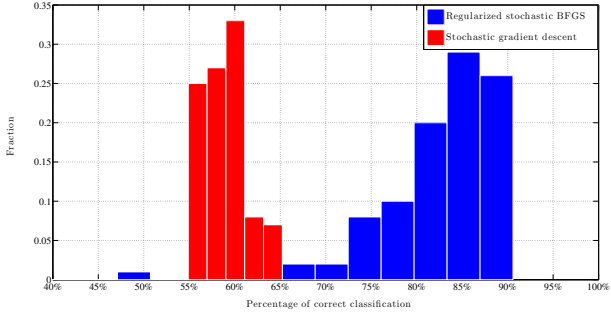}
\caption{Histogram of correct classification percentages for dimension $n=4$ and training set of size $N=2.5\times 10^3$. Vectors computed by RES outperform those computed via SGD and are not far from the accuracy of clairvoyant classifiers (test sets contain $10^4$ samples; histogram is across $10^3$ realizations; parameters as in Fig. \ref{fig:n=4}).}
\label{accuracy}
\end{figure}

\begin{figure}[t]
\centering
\includegraphics[width=\linewidth,height=0.48\linewidth]{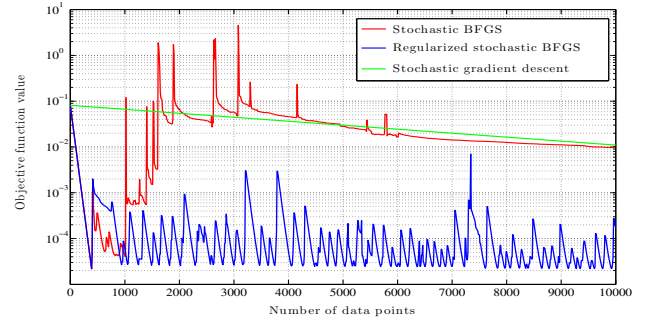}
\caption{Comparison of SGD, regularized stochastic BFGS (RES), and (non regularized) stochastic BFGS. The regularization is fundamental to control the erratic behavior of stochastic BFGS (sample size $L=5$; constant stepsize $\epsilon_{t} = 10^{-1}$; Regularization parameters  $\delta=10^{-3}$ and $\Gamma=10^{-4}$, feature vector dimension $n=10$).}
\label{jumps}
\end{figure}

\section{Conclusions}

Convex optimization problems with stochastic objectives were considered. RES, a stochastic implementation of a regularized version of the Broyden-Fletcher-Goldfarb-Shanno quasi-Newton method was introduced to find corresponding optimal arguments. Almost sure convergence was established under the assumption that sample functions have well behaved Hessians. A linear convergence rate in expectation was further proven. Numerical results showed that RES affords important reductions in terms of convergence time relative to stochastic gradient descent. These reductions are of particular significance for problems with large condition numbers or large dimensionality since RES exhibits remarkable stability in terms of the total number of iterations required to achieve target accuracies. An application of RES to support vector machines was also developed. In this particular case the advantages of RES manifest in improvements of classification accuracies for training sets of fixed cardinality. Future research directions include the development of limited memory versions as well as distributed versions where the function to be minimized is spread over agents of a network.

\section*{Appendix A: Proof of Proposition \ref{flen}}

%
We first show that \eqref{eqn_inverse_appx_update} is true. Since the optimization problem in \eqref{jadid} is convex in $\bbZ$ we can determine the optimal variable $\bbB_{t+1}=\bbZ^{*}$ using Lagrangian duality. Introduce then the multiplier variable $\bbmu$ associated with the secant constraint $\bbZ \bbv_{t} =  \bbr_{t}$ in  \eqref{jadid} and define the Lagrangian
\begin{align}\label{bacon}
   \ccalL(\bbZ,\bbmu)\ = & \ 
       \tr(\bbB_{t}^{-1}(\bbZ-\delta \bbI))  
       - \log \det(\bbB_{t}^{-1}(\bbZ-\delta \bbI)) - n \nonumber \\ & \qquad
       + \bbmu^{T} \left(\bbZ \bbv_{t}- \bbr_{t}\right).
\end{align}
The dual function is defined as $d(\bbmu):=\min_{\bbZ \succeq\bbzero}\ccalL(\bbZ,\bbmu)$ and the optimal dual variable is $\bbmu^* := \argmin_{\bbmu} d(\bbmu)$. We further define the primal Lagrangian minimizer associated with dual variable $\bbmu$ as 
\begin{equation}\label{star}
    \bbZ (\bbmu) := \argmin_{\bbZ \succeq\bbzero} \ccalL(\bbZ,\bbmu) .
\end{equation}
Observe that combining the definitions in \eqref{star} and \eqref{bacon} we can write the dual function $d(\bbmu)$ as
\begin{alignat}{2}\label{eqn_lemma_flen_pf_10}
d(\bbmu) 
   \ =\ &\ \ccalL(\bbZ(\bbmu),\bbmu) \nonumber \\ 
   \ =\ 
         &\ \tr(\bbB_{t}^{-1}(\bbZ(\bbmu)-\delta \bbI))  
         - \log \det(\bbB_{t}^{-1}(\bbZ(\bbmu)-\delta \bbI))  \nonumber \\  & \qquad
         - n + \bbmu^{T} \left(\bbZ(\bbmu) \bbv_{t}- \bbr_{t}\right).
\end{alignat}
We will determine the optimal Hessian approximation $\bbZ^{*}=\bbZ(\bbmu^*)$ as the Lagrangian minimizer associated with the optimal dual variable $\bbmu^{*}$. To do so we first find the Lagrangian minimizer \eqref{star} by nulling the gradient of $\ccalL(\bbZ,\bbmu)$ with respect to $\bbZ$ in order to show that $\bbZ(\bbmu)$ must satisfy
\begin{equation}\label{popolov}
   \bbB_{t}^{-1} 
     - \left(\bbZ (\bbmu) -\delta \bbI\right)^{-1} 
     + \frac{\bbmu \bbv_{t}^{T} + \bbv_{t}  \bbmu^{T}}{2}
   = 0  .
\end{equation}
Multiplying the equality in \eqref{popolov} by $\bbB_{t}$ from the right and rearranging terms  it follows that the inverse of the argument of the log-determinant function in \eqref{eqn_lemma_flen_pf_10} can be written as
\begin{equation}\label{shiraz}
(\bbZ (\bbmu)-\delta \bbI) ^{-1}\bbB_{t}= \bbI+\left({{\bbmu \bbv_{t}^{T} + \bbv_{t}  \bbmu^{T}}\over{2}} \right)\bbB_{t}.
\end{equation}
If, instead, we multiply \eqref{popolov} by $(\bbZ (\bbmu)-\delta \bbI)$  from the right it follows after rearranging terms that
\begin{equation}\label{juventus}
   \bbB_{t}^{-1}(\bbZ (\bbmu)-\delta \bbI) 
       =  \bbI -  {{\bbmu \bbv_{t}^{T} + \bbv_{t}  \bbmu^{T}}\over{2}}  (\bbZ (\bbmu)-\delta \bbI).
\end{equation}
Further considering the trace of both sides of \eqref{juventus} and noting that $\tr(\bbI)=n$ we can write the trace in \eqref{eqn_lemma_flen_pf_10} as
\begin{equation}\label{milan}
    \tr(\bbB_{t}^{-1}(\bbZ (\bbmu)-\delta \bbI))=n - \tr\Big{[}{{\bbmu \bbv_{t}^{T} + \bbv_{t}  \bbmu^{T}}\over{2}}  (\bbZ (\bbmu)-\delta \bbI)\Big{]} .
\end{equation}
Observe now that since the trace of a product is invariant under cyclic permutations of its arguments and the matrix $\bbZ$ is symmetric we have $\tr[\bbmu\bbv_{t}^{T}(\bbZ (\bbmu)-\delta \bbI)]=\tr[\bbv\bbmu_{t}^{T}(\bbZ (\bbmu)-\delta \bbI)]=\tr[\bbmu^{T}(\bbZ (\bbmu)-\delta \bbI)\bbv_{t}]$. Since the argument in the latter is a scalar the trace operation is inconsequential from where it follows that we can rewrite \eqref{milan} as
 \begin{equation}\label{chelsea}
\tr(\bbB_{t}^{-1}(\bbZ (\bbmu)-\delta \bbI))=n - \bbmu^{T}(\bbZ (\bbmu)-\delta \bbI)\bbv_{t}.
\end{equation}
Observing that the log-determinant of a matrix is the opposite of the log-determinant of its inverse we can substitute \eqref{shiraz} for the argument of the log-determinant in \eqref{eqn_lemma_flen_pf_10}. Further substituting \eqref{chelsea} for the trace in \eqref{eqn_lemma_flen_pf_10} and rearranging terms yields the explicit expression for the dual function
\begin{equation}\label{marlboro}
  d(\bbmu)=\log \det \Bigg{[}\bbI+\Big{(}{{\bbmu \bbv_{t}^{T} + \bbv_{t}  \bbmu^{T}}\over{2}} \Big{)} \bbB_{t} \Bigg{]}-\bbmu^{T} (\bbr_{t}-\delta \bbv_{t}).
\end{equation}
In order to compute the optimal dual variable $\bbmu^{*}$ we set the gradient of \eqref{marlboro} to zero and manipulate terms to obtain  
\begin{equation}\label{opel}
\bbmu^{*} = {{1 }\over{ {{}{\tbr_{t}^{T }\bbv_{t}}} }}  {{ \Bigg{(} \bbv_{t} \bigg{(} 1+ {{\tbr_{t}^{T}\bbB_{t}^{-1}\tbr_{t}}\over{ \tbr_{t}^{T }\bbv_{t}}}\bigg{)} - 2 \bbB_{t}^{-1} \tbr_{t} \Bigg{)} }},
\end{equation}
where we have used the definition of the corrected gradient variation $\tbr_{t} := \bbr_{t}-\delta \bbv_{t}$. To complete the derivation plug the expression for the optimal multiplier $\bbmu^{*}$ in \eqref{opel} into the Lagrangian minimizer expression in \eqref{popolov} and regroup terms so as to write
\begin{equation}\label{esteghlal}
(\bbZ(\bbmu^{*})-\delta \bbI)^{-1}={{\bbv_{t} \bbv_{t}^{T}}\over{\tbr_{t}^{T }\bbv_{t}}} +\left(\bbI-{{\bbv_{t} \tbr_{t}^{T}}\over{\tbr_{t}^{T }\bbv_{t}}}\right)\bbB_{t}^{-1}\left(\bbI-{{\tbr_{t} \bbv_{t}^{T}}\over{\tbr_{t}^{T } \bbv_{t}}}\right).
\end{equation}
Applying the Sherman-Morrison formula to compute the inverse of the right hand side of \eqref{esteghlal} leads to 
\begin{equation}\label{ghahreman}
\bbZ(\bbmu^{*}) - \delta \bbI=\bbB_{t} + {{ \tbr_t  \tbr_t^{T}}\over{\bbv_{t}^{T} \tbr_t}}- {{\bbB_{t} \bbv_{t}\bbv_{t}^{T}{\bbB_{t}} }\over{\bbv_{t}^{T}\bbB_{t}\bbv_{t}}}, 
\end{equation}
which can be verified by direct multiplication. The result in \eqref{eqn_inverse_appx_update} follows after solving \eqref{ghahreman} for $\bbZ(\bbmu^{*})$ and noting that for the convex optimization problem in \eqref{jadid} we must have $\bbZ(\bbmu^{*})=\bbZ^{*}=\bbB_{t+1}$ as we already argued.

To prove \eqref{min_eigenvalue_condition} we operate directly from \eqref{eqn_inverse_appx_update}. Consider first the term $\tbr_{t} \tbr_{t}^{T}/\bbv_{t}^{T} \tbr_{t}$ and observe that since the hypotheses include the condition $\bbv_{t}^{T} \tbr_{t}>0$, we must have
\begin{equation}\label{first_term}
\frac{\tbr_{t} \tbr_{t}^{T}}{\bbv_{t}^{T} \tbr_{t}}\ \succeq\ \bb0.
\end{equation}
Consider now the term $\bbB_{t}- {{\bbB_{t} \bbv_{t}\bbv_{t}^{T}{\bbB_{t}} }/{\bbv_{t}^{T}\bbB_{t}\bbv_{t}}} $ and factorize $\bbB_{t}^{{1}/{2}}$ from the left and right side so as to write
\begin{equation}\label{second_term_prelim}
\bbB_{t} - \frac{{\bbB_{t} \bbv_{t}\bbv_{t}^{T}{\bbB_{t}} }}{{\bbv_{t}^{T}\bbB_{t}\bbv_{t}}}\  =\ \bbB_{t}^{1/2}
		 \left(   \bbI -  \frac{{\bbB_{t}^{1/2} \bbv_{t}\bbv_{t}^{T}{\bbB_{t}^{1/2}} }}{{\bbv_{t}^{T}\bbB_{t}\bbv_{t}}}  \right) \bbB_{t}^{1/2}
\end{equation}
Define the vector $\bbu_t:=\bbB_{t}^{1/2}\bbv_{t}$ and write $\bbv_{t}^{T}\bbB_{t}\bbv_{t} = (\bbB_{t}^{1/2}\bbv_{t})^T(\bbB_{t}^{1/2}\bbv_{t}) = \bbu_t^T\bbu_t$ as well as $\bbB_{t}^{1/2} \bbv_{t}\bbv_{t}^{T}\bbB_{t}^{1/2} = \bbu_t\bbu_t^T$. Substituting these observation into \eqref{second_term_prelim} we can conclude that 
\begin{equation}\label{second_term}
\bbB_{t}\ -\ \frac{{\bbB_{t} \bbv_{t}\bbv_{t}^{T}{\bbB_{t}} }}{{\bbv_{t}^{T}\bbB_{t}\bbv_{t}}}\  =\ \bbB_{t}^{1/2}
		 \left(   \bbI -  \frac{\bbu_t\bbu_t^T}{\bbu_t^T\bbu_t}\right)\bbB_{t}^{1/2} \succeq\ \bb0,
\end{equation}
because the eigenvalues of the matrix $\bbu_t\bbu_t^T/\bbu_t^T\bbu_t$ belong to the interval $[0,1]$. The only term in \eqref{eqn_inverse_appx_update} which has not been considered is $\delta\bbI$. Since the rest add up to a positive semidefinite matrix it then must be that \eqref{min_eigenvalue_condition} is true.


\section*{Appendix B: Proof of Theorem \ref{theo_convergence_rate}}

Theorem \ref{theo_convergence_rate} claims that the sequence of expected objective values $\E {F(\bbw_{t})}$ approaches the optimal objective $F(\bbw^*)$ at a linear rate $O(1/t)$. Before proceeding to the proof of Theorem \ref{theo_convergence_rate} we introduce a technical lemma that provides a sufficient condition for a sequence $u_t$ to exhibit a linear convergence rate.

%
\begin{lemma}\label{lecce22}
Let $c>1$, $b>0$ and $t_0 > 0$ be given constants and $u_{t}\geq 0$ be a nonnegative sequence that satisfies the inequality
\begin{equation}\label{claim23}
   u_{t+1} \leq \left( 1- \frac{c}{t+t_0} \right) u_{t} 
                 + \frac{b}{{(t+t_0)}^{2}}\ ,
\end{equation}
{for all times $t\geq0$}. The sequence $u_t$ is then bounded as
\begin{equation}\label{lemma3_claim}
u_{t} \leq\  \frac{Q}{t+t_{0}},
\end{equation}
for all times $t\geq0$, where the constant $Q$ is defined as
\begin{equation}\label{convergence_parameter}
   Q:=\max \left[\frac{b}{c-1},\ t_{0} u_{0}  \right] .
\end{equation}
\end{lemma}

%
\begin{myproof}
We prove \eqref{lemma3_claim} using induction. To prove the claim for {$t=0$} simply observe that the definition of $Q$ in \eqref{convergence_parameter} implies that
\begin{equation}\label{eqn_lecce_pf_10}
   Q:=\max \left[\frac{b}{c-1},\ t_{0} u_{0}  \right] \geq\ t_{0} u_{0},
\end{equation}
because the maximum of two numbers is at least equal to both of them. By rearranging the terms in \eqref{eqn_lecce_pf_10} we can conclude that 
\begin{equation}\label{first_step_of_induction}
    u_0\ \leq\  \frac{Q}{t_{0}}.
\end{equation}
Comparing \eqref{first_step_of_induction} and \eqref{lemma3_claim} it follows that the latter inequality is true for $t=0$.

Introduce now the induction hypothesis that \eqref{lemma3_claim} is true for $t=s$. To show that this implies that \eqref{lemma3_claim} is also true for $t=s+1$ substitute the induction hypothesis $u_s\leq Q/(s+t_0)$ into the recursive relationship in \eqref{claim23}. This substitution shows that $u_{s+1}$ is bounded as
\begin{equation}\label{hassan_hassan}
    u_{s+1} \leq 
        \left( 1- \frac{c}{s+t_0} \right) \frac{Q}{s+t_{0}}
        +\frac{b}{{(s+t_0)}^{2}} \ .
\end{equation}
Observe now that according to the definition of $Q$ in \eqref{convergence_parameter}, we know that $b/(c-1) \leq Q$ because $Q$ is the maximum of $b/(c-1)$ and $t_{0}u_{0}$. Reorder this bound to show that $b\leq Q(c-1)$ and substitute into \eqref{hassan_hassan} to write
\begin{align}\label{eqn_lecce_pf_40}
   u_{s+1} \leq 
        \left( 1- \frac{c}{s+t_0} \right) \frac{Q}{s+t_{0}}  
        +\frac{(c-1) Q}{{(s+t_0)}^{2}}  \ .
\end{align}
Pulling out $Q/(s + t_0)^2$ as a common factor and simplifying and reordering terms it follows that \eqref{eqn_lecce_pf_40} is equivalent to
\begin{align}\label{jhn}
   u_{s+1} 
       \ \leq\  \frac{Q \big[s+t_0 -c +(c-1)\big]}{{(s+t_0)}^{2}} 
       \ =   \  \frac{s+t_0-1}{{(s+t_0)}^{2}}  Q .
\end{align}
To complete the induction step use the difference of squares formula for $(s+t_0)^2 - 1$ to conclude that 
\begin{equation}\label{algebra}
   \big[ (s+t_0) - 1\big]     \big[ (s+t_0) + 1\big]  
       \ =   \ (s+t_0)^2 - 1 
       \ \leq\ (s+t_0)^2.
\end{equation}
Reordering terms in \eqref{algebra} it follows that $\big[ (s+t_0) - 1\big]/ (s+t_0)^2 \leq 1/\big[ (s+t_0) + 1\big]$, which upon substitution into \eqref{jhn} leads to the conclusion that 
\begin{align}\label{eqn_lecce_pf_80}
    u_{s+1} \leq  \frac{Q}{s+t_0 + 1}.
\end{align}
Eq. \eqref{eqn_lecce_pf_80} implies that the assumed validity of \eqref{lemma3_claim} for $t=s$ implies the validity of \eqref{lemma3_claim} for $t=s+1$. Combined with the validity of \eqref{lemma3_claim} for $t=0$, which was already proved, it follows that \eqref{lemma3_claim} is true for all times $t\geq0$.
 \end{myproof}

%
Lemma \ref{lecce22} shows that satisfying \eqref{claim23} is sufficient for a sequence to have the linear rate of convergence specified in \eqref{lemma3_claim}. In the following proof of Theorem \ref{theo_convergence_rate} we show that if the stepsize sequence parameters $\epsilon_{0}$ and $T_{0}$ satisfy \eqref{eqn_thm_cvg_rate_10} the sequence $\E{F(\bbw_{t})}-F(\bbw^*)$ of expected optimality gaps satisfies \eqref{claim23} with {$c=2\epsilon_{0}T_{0}\Gamma$, $b=\epsilon_{0}^{2} T_{0}^{2} K$ and $t_0=T_{0}$.} The result in \eqref{eqn_thm_cvg_rate_20} then follows as a direct consequence of Lemma \ref{lecce22}.

%
\medskip \noindent {\bf Proof of Theorem \ref{theo_convergence_rate}: }
Consider the result in \eqref{pedarsag} of Lemma \ref{helpful} and subtract the average function optimal value $F(\bbw^*)$ from both sides of the inequality to conclude that the sequence of optimality gaps in the RES algorithm satisfies
\begin{align}\label{taylor_objective}
 &\E{F(\bbw_{t+1})\given \bbw_{t}} -\ F(\bbw^*)   \\ \nonumber &\qquad\qquad\qquad
         \leq F(\bbw_{t}) -\ F(\bbw^*) 
          -  \epsilon_{t} \Gamma \| \nabla F(\bbw_{t})\|^{2} 
          +\epsilon_{t}^{2} K,
\end{align}
where, we recall, $K:={MS^{2}} (({1/\delta})+\Gamma)^{2} /2$ by definition.

We proceed to find a lower bound for the gradient norm $\| \nabla F(\bbw_{t})\|$ in terms of the error of the objective value $F(\bbw_{t}) -\ F(\bbw^*) $ {-- this is a standard derivation which we include for completeness, see, e.g., \cite{Boyd}. As it follows from Assumption 1 the eigenvalues of the Hessian $\bbH(\bbw_{t})$ are bounded between $0<m$ and $M<\infty$ as stated in \eqref{bbb}. Taking a Taylor's expansion of the objective function $F(\bby)$ around $\bbw$ and using the lower bound in the Hessian eigenvalues we can write
\begin{equation}\label{taylor_lower_bound}
   F(\bby) \geq\ F(\bbw) +\nabla F(\bbw)^{T}(\bby-\bbw)   
   + {{m}\over{2}}\|{\bby - \bbw}\|^{2}.
\end{equation}
For fixed $\bbw$, the right hand side of \eqref{taylor_lower_bound} is a quadratic function of $\bby$ whose minimum argument we can find by setting its gradient to zero. Doing this yields the minimizing argument $\hby = \bbw- (1/m) \nabla  F(\bbw)$ implying that for all $\bby$ we must have
\begin{alignat}{2}\label{lower_bound_for_gradient}
F(\bby) \geq\ 
    &\ F(\bbw) +\nabla F(\bbw)^{T}(\hby-\bbw)   
   + {{m}\over{2}}\|{\hby - \bbw}\|^{2} \nonumber \\
   \ =\ 
         &\ F(\bbw) - \frac{1}{2m} \| \nabla F(\bbw)\|^{2} .
\end{alignat}
The bound in \eqref{lower_bound_for_gradient} is true for all $\bbw$ and $\bby$. In particular, for $\bby=\bbw^{*}$ and $\bbw=\bbw_{t}$ \eqref{lower_bound_for_gradient} yields
\begin{equation}\label{transition}
   F(\bbw^*) \geq  F(\bbw_{t}) - \frac{1}{2m} \| \nabla F(\bbw_{t})\|^{2}.
\end{equation} 
Rearrange terms in \eqref{transition} to obtain a bound on the gradient norm squared $\| \nabla F(\bbw_{t})\|^{2}$. Further substitute the result in \eqref{taylor_objective} and regroup terms to obtain the bound
\begin{align}\label{simplified_taylor_expansion}
 &\E{F(\bbw_{t+1})\given \bbw_{t}} -\ F(\bbw^*)   \\ \nonumber &\qquad\qquad
         \leq\ (1-2m\epsilon_{t} \Gamma) \big{(}F(\bbw_{t}) -\ F(\bbw^*) \big{)} 
          +\epsilon_{t}^{2} K .
\end{align}
Take now expected values on both sides of \eqref{simplified_taylor_expansion}. The resulting double expectation in the left hand side simplifies to $\E{  \E {F(\bbw_{t+1})\given{\bbw_{t}}} } = \E {F(\bbw_{t+1})}$, which allow us to conclude that \eqref{simplified_taylor_expansion} implies that
\begin{align}\label{expectation_inequality}
 &  \E{F(\bbw_{t+1})} -\ F(\bbw^*)     \\ \nonumber &\qquad\qquad
         \leq\ (1-2m\epsilon_{t} \Gamma) \big{(} \E {F(\bbw_{t})} -\ F(\bbw^*) \big{)}
          +\epsilon_{t}^{2} K   .
\end{align}
Furhter substituting $\epsilon_{t} \!= \! \epsilon_{0}T_{0}/(T_{0}+t)$, which is the assumed form of the step size sequence by hypothesis, we can rewrite \eqref{expectation_inequality} as
\begin{align}\label{sass}
& \E{F(\bbw_{t+1})} -\ F(\bbw^*)  \\ \nonumber &  \quad \!\!
        \leq \left(1- \frac{2\ \epsilon_{0}T_{0} \Gamma}{(T_{0}+t)} \right) 
                     \Big( \E {F(\bbw_{t})} - F(\bbw^*) \Big)
             +\left(\frac{\epsilon_{0}T_{0}}{T_{0}+t}\right)^{2} \! K.
\end{align}
Given that the product $2 \epsilon_{0} T_{0} \Gamma >1$ as per the hypothesis in \eqref{eqn_thm_cvg_rate_10} the sequence $\E{F(\bbw_{t+1})} -\ F(\bbw^*)$ satisfies the hypotheses of Lemma \ref{lecce22} with {$c=2\epsilon_{0}T_{0}\Gamma$, $b=\epsilon_{0}^{2} T_{0}^{2} K$ and $t_0=T_{0}$.} It then follows from \eqref{lemma3_claim} and \eqref{convergence_parameter} that \eqref{eqn_thm_cvg_rate_20} is true for the $C_{0}$ constant defined in \eqref{eqn_thm_cvg_rate_30} upon identifying $u_t$ with $\E{F(\bbx_{t+1})} -\ F(\bbx^*)$, $C_{0}$ with $Q$, and substituting {$c=2\epsilon_{0}T_{0}\Gamma$, $b=\epsilon_{0}^{2} T_{0}^{2} K$ and $t_0=T_{0}$} for their explicit values.

\bibliographystyle{IEEEtran}
  \bibliography{01_my_journals_2005_2008,bib_files/02_my_conferences_2009_2010,bib_files/02_my_conferences_2004_2008,bib_files/01_my_journals_2005_2008,bib_files/01_my_journals_2009_2010,bib_files/02_my_conferences_2011,bib_files/02_my_conferences_2012,bib_files/01_my_journals_2012,bib_files/01_my_journals_2011,bib_files/04_bib_cross_layer,bib_files/yichuan_1,bib_files/yichuan_2}
   \end{document}